\documentclass[fleqn,10pt]{wlscirep}
\usepackage[utf8]{inputenc}
\usepackage[T1]{fontenc}
\usepackage{lineno}
\usepackage{float}
\usepackage{array}
\usepackage{color}
\usepackage{colortbl}
\definecolor{orange}{rgb}{1,0.647,0}

\newcolumntype{P}[1]{>{\centering\arraybackslash}p{#1}}
\newcolumntype{M}[1]{>{\centering\arraybackslash}m{#1}}

\title{The Muon Space GNSS-R Surface Soil Moisture Product}  

\author[1,$\dag$*]{Max Roberts}
\author[1,$\dag$]{Ian Colwell}
\author[1,$\dag$]{Clara Chew}
\author[1]{Dallas Masters}
\author[1]{Karl Nordstrom}
\affil[1]{Muon Space, Science Team, 2250 Charleston Rd, Mountain View, CA 94043}

\affil[*]{corresponding author: Max Roberts (max@muonspace.com)}

\affil[$\dag$]{these authors contributed equally to this work}

\begin{abstract}

Muon Space (Muon) is building a constellation of small satellites, many of which will carry global navigation satellite system-reflectometry (GNSS-R) receivers. In preparation for the launch of this constellation, we have developed a generalized deep learning retrieval pipeline, which now produces operational GNSS-R near-surface soil moisture retrievals using data from NASA's Cyclone GNSS (CYGNSS) mission. In this article, we describe the input datasets, preprocessing methods, model architecture, development methods, and detail the soil moisture products generated from these retrievals. The performance of this product is quantified against in situ measurements and compared to both the target dataset (retrievals from the Soil Moisture Active-Passive (SMAP) satellite) and the v1.0 soil moisture product from the CYGNSS mission. The Muon Space product achieves improvements in spatial resolution over SMAP with comparable performance in many regions. An ubRMSE of 0.032 cm3 cm-3 for in situ soil moisture observations from SMAP core validation sites is shown, though performance is lower than SMAP's when comparing in forests and/or mountainous terrain. The Muon Space product outperforms the v1.0 CYGNSS soil moisture product in almost all aspects. This initial release serves as the foundation of our operational soil moisture product, which soon will additionally include data from Muon Space satellites.

\end{abstract}
\begin{document}

\flushbottom
\maketitle

\thispagestyle{empty}

\section*{Background \& Summary}

Muon Space ("Muon") is building a constellation of small satellites to collect Earth observations for commercial, government, and non-governmental organization (NGO) customers, focusing on monitoring and addressing issues related to climate change as well as weather applications. Remote sensing payloads and products derived from global navigation satellite system (GNSS) signals are some of the first being developed by Muon. Advancement in the quantity and maturity of GNSS reflectometry (GNSS-R) data products has expanded since the launch of the Surrey Satellite Systems TechDemoSat-1 (TDS-1), NASA Cyclone GNSS (CYGNSS), and some initial commercial missions. The L-band GNSS signals can penetrate moderate vegetation cover as well as the top 5 cm of the soil surface, making this type of data well-suited for retrieving soil moisture (SM), among other terrestrial and oceanic applications. In preparation for GNSS-R data from Muon satellites, we have developed and operationalized a soil moisture product that utilizes data from CYGNSS as input, which is the focus of this article. As Muon's GNSS-R constellation grows, we will add to the 8+ years of GNSS-R data that are already available, improving the revisit rate and spatial coverage, and creating a long-term SM record that will extend into the future beyond the lifetime of CYGNSS.

\begin{figure}[H]
    \centering
    \includegraphics[width=\linewidth]{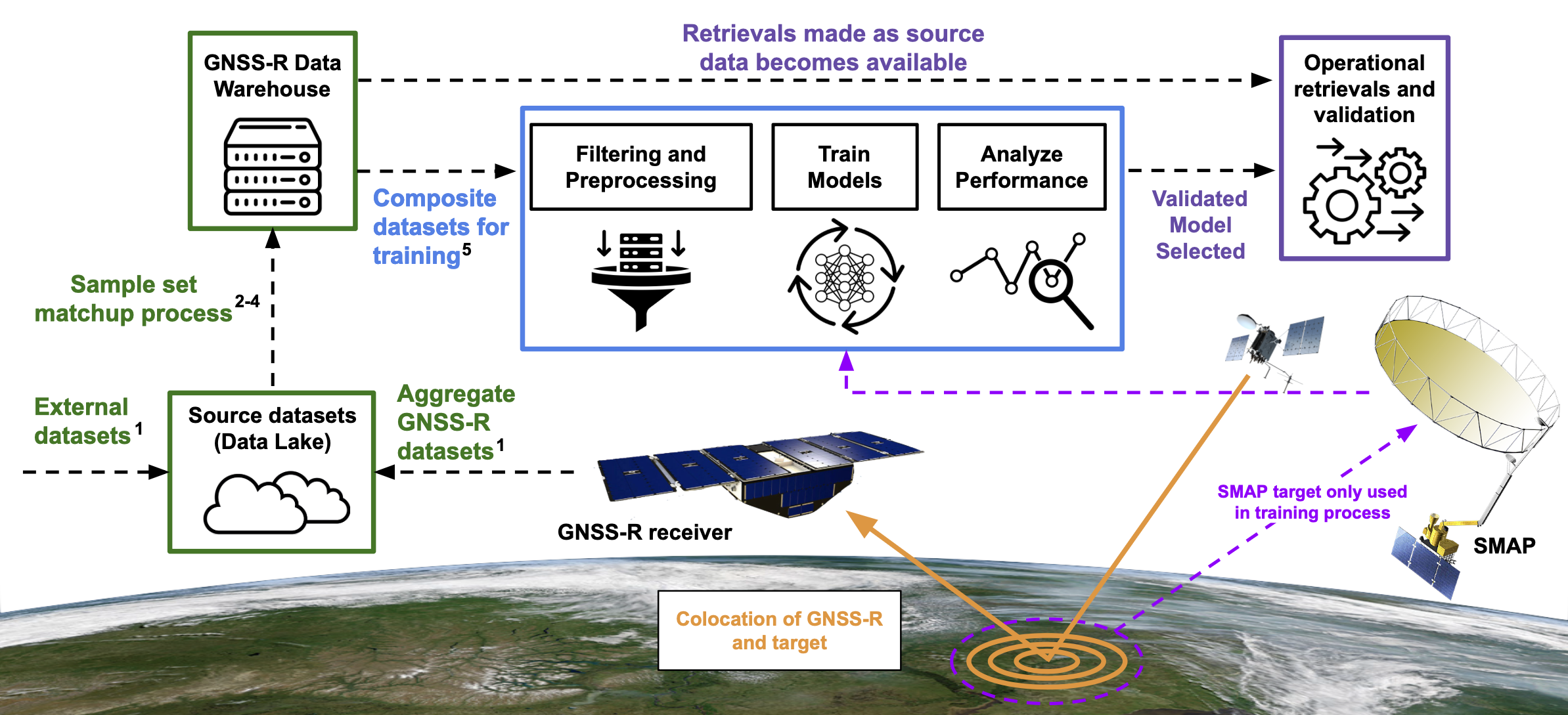}
    \caption{An illustration of the application of the Muon generalized GNSS-R retrieval pipeline to SM, which transforms L1 GNSS-R land surface observations into operational L2 and L3 data products. The green boxes and text represent the data pipeline components, while the blue box and text represent the model development process. The dark purple box and text shows operational product generation using a finalized DL model. The CYGNSS and SMAP satellites depict the matching of target information (i.e., “truth”) to each GNSS reflection measurement.}
    \label{fig:retrieval_flow}
\end{figure}

Empirical and semi-empirical methods\cite{chew_description_2020, kim_use_2018} and deep learning (DL) techniques\cite{eroglu_high_2019, senyurek_evaluations_2020, roberts_deep-learning_2022} have successfully demonstrated the retrieval of SM from GNSS-R data. Building on the work of Roberts et al.\cite{roberts_deep-learning_2022}, we have developed Level 2 (L2) along-track and Level 3 (L3) gridded GNSS-R SM products. Roberts et al.\cite{roberts_deep-learning_2022} used convolutional neural networks (CNNs) to develop a data-driven model relating the GNSS reflection measurement to surface parameters and laid the groundwork for a mechanism to achieve improved GNSS-R SM retrievals (illustrated in Figure \ref{fig:retrieval_flow}). Here, the model was trained with CYGNSS bistatic radar delay-Doppler maps (DDMs) and contextual ancillary datasets as inputs along with a selected target (i.e., "truth") SM. Although any source of coincident SM could be used as a target, retrievals from the NASA Soil Moisture Active-Passive (SMAP) mission were selected since they spatially and temporally overlap with CYGNSS observations. Additionally, forward-scattered L-band GNSS-R measurements fundamentally observe and are affected by similar phenomenon as L-band radiometers, such as SMAP. Validation of the SM retrievals against in situ data shows similar performance to the SMAP retrievals in low- to moderately-vegetated environments, though with improved spatial resolution. Densely forested and mountainous regions currently underperform SMAP, and improving performance in these areas is a key focus of Muon's ongoing research. The initial GNSS-R SM product based on CYGNSS v3.2 data described here are produced operationally and are made available to the public. Community utilization, feedback, and adoption are the primary objectives of releasing this CYGNSS-based dataset. The data preprocessing and training algorithms developed using CYGNSS data will subsequently be applied to the Muon GNSS-R datasets as they become available in late 2024, with the goal of creating a long-term, operational GNSS-R SM product from harmonized data derived multiple satellite missions.

\section*{Methods}

We start our discussion by motivating the use of DL methods for GNSS-R retrievals and then describe the generalized processing pipeline developed to ingest GNSS-R observations and ancillary input datasets, train a model using SMAP SM as a target, and use the model to make SM retrievals from GNSS-R observations. This includes detailing the steps involved in dataset filtering and preprocessing and our strategies for model development. We then outline our model validation process and review our methods to quantify feature importance, input justification, model explainability, and uncertainty quantification. Finally, we discuss the additional processes to generate L2 and L3 datasets from our model.

\subsection*{Deep Learning for Geophysical Retrievals from GNSS-R Delay-Doppler Maps}

The bistatic scattering of L-band GNSS signals from land surfaces and the resulting DDMs of reflected signal power across delay and Doppler are difficult to model. While a number of analytical models can simulate reflections under specific conditions such as a simple air-soil interface or including attenuation by a vegetation layer\cite{SAVERS}, a complete representation of the surface scattering is not tractable for the complex scattering environments for realistic surface conditions. There are many factors that affect the features observed in DDMs, including the soil type, surface roughness, overlying vegetation, spacecraft bistatic geometry, and the specific GNSS signal characteristics. Traditionally, scalar observables, such as an average of the DDM samples surrounding the nominal specular reflection point delay or the slope of the leading edge of the delay waveform, are derived from the DDM and are used to construct data-driven retrieval models. Retrievals of wind speed from DDMs generated by reflections from the ocean surface (a mostly homogeneous dielectric surface) use multivariate regression of these scalars against a target dataset\cite{clarizia_wind_2016}, and similarly, retrievals of SM use regression-based approaches based on a derived reflectivity from the peak power of the DDM\cite{chew_description_2020}. These approaches often compress or ignore a significant amount of potential information contained in the DDM. Methods built on DL present an opportunity to construct models that ingest the entire DDM as one input, as well as ancillary contextual information about the reflecting surface. For SM, this enables the inclusion of information such as soil texture, vegetation, and the presence of surface water in the field of view (FOV), all of which affect the scattering of L-band signals and the resulting observed DDM. This additional context presents an advantage over traditional analytic/empirical methods\cite{chew_description_2020}, where including this information is more challenging.

\subsection*{Generalized GNSS-R Retrieval Pipeline}

As discussed above, DL models offer benefits in retrieving geophysical parameters from DDMs collected from signals scattered by complex, heterogeneous surfaces. Using this same architecture for a variety of retrieval types, including simpler ocean scattering regimes for retrieving ocean wind speed, allows higher-dimensional relationships than are possible with traditional retrieval techniques. Additionally, having a single, unified architecture for the interpretation of GNSS-R measurements allows for the development of multiple products with a degree of parallelization and code reuse, where research and development into one retrieval type often benefits other retrievals. 

Therefore, we have developed a generalized approach to performing retrievals from GNSS-R measurements built on a DL analysis framework. A visualization of this pipeline is shown in Figure \ref{fig:retrieval_flow} for the application to SM. The data ingestion pipeline elements are represented in green, including the aggregation of source datasets in a data lake, and the restructuring of these datasets specific for our applications, referred to as the "data warehouse." The model development elements, which include preprocessing of the data, training the model, and interpretation of model performance are illustrated in blue. The operational elements that stream low-level GNSS-R observations to trained models to generate geophysical retrievals of higher level products is shown in dark purple. An automated validation process to monitor deviations in the model performance in real-time is currently under development. The entire pipeline is implemented using cloud-based services (e.g., Amazon Web Services) and has been optimized for object storage, access, and processing on these platforms. We make note of the generalizablity of this pipeline, but the rest of the article is focused specifically on the application to SM.

\subsection*{Source Dataset Selection}

\subsubsection*{Target Data: SMAP Soil Moisture}

The SMAP Enhanced L3 Radiometer Global and Polar Grid Daily 9 km EASE-Grid Soil Moisture (Version 5)\cite{oneill_peggy_e_smap_2021} represents the best available soil moisture "truth" dataset spanning the CYGNSS data record and geographical sampling region. SMAP descending pass (AM) retrievals that are collocated with a CYGNSS observation on the same calendar day create a potential training "sample." An example of a single day of SMAP AM coverage is shown in Figure \ref{fig:coverage}, highlighting that there is a significant fraction of the planet not used for training on any given day. Note that while SMAP ascending passes (PM) data have been historically ignored due to potential issues with thermal gradients in soil and canopies\cite{jackson_soil_2017}, here they were omitted because the SMAP AM data was sufficient for training purposes.

\subsubsection*{Primary Input: CYGNSS GNSS-R Delay-Doppler Maps}

\begin{figure}[H]
    \centering
    \includegraphics[width=1\linewidth]{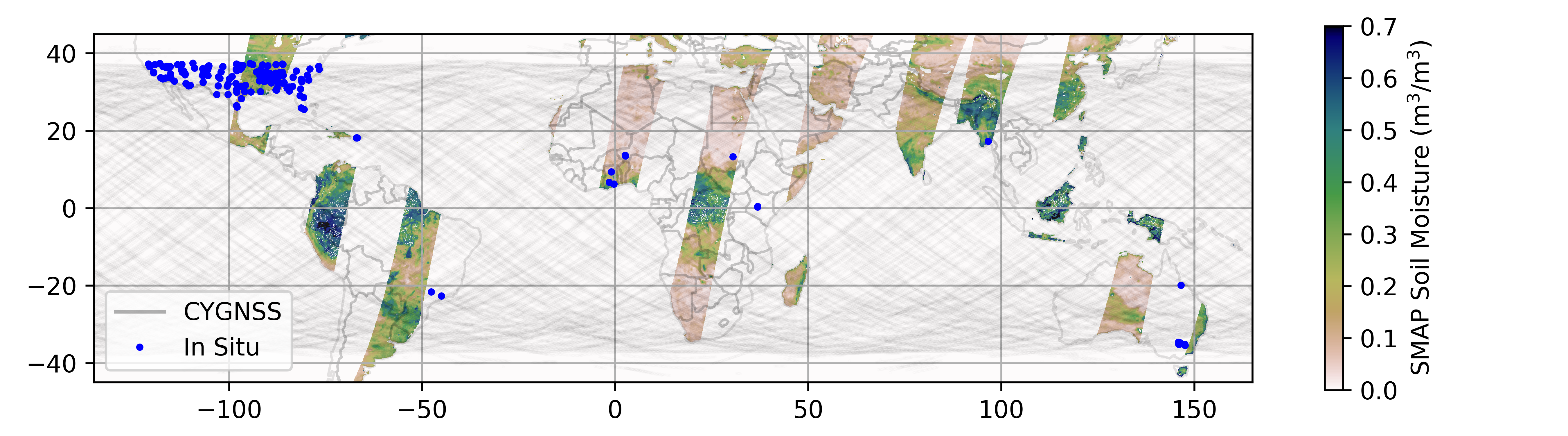}
    \caption{Observation coverage from one day (May 18, 2022) sampled by the CYGNSS constellation (grey tracks). Also shown is the SMAP Level 3 gridded SM product showing AM coverage (colored swaths), as well as the locations of the in situ validation sites (blue dots). CYGNSS data are used as the primary model input, SMAP AM data are used as the target for training, and the in situ data are used for validation of the model output.}
    \label{fig:coverage}
\end{figure}

The primary input data are DDMs from the CYGNSS Level 1 Science Data Record Version 3.2\cite{cygnss_L1}. Due to the bistatic geometry of GNSS transmitters, the reflecting Earth surface, and low Earth orbit (LEO) satellite receivers, spaceborne GNSS-R observations form "tracks" that are pseudorandomly distributed across the Earth surface. These tracks are produced along the path of the specular reflection point defined as the point on the surface where the angles between the the incident and reflected rays are equal. Tracks from one day of collections from the eight small satellites comprising the CYGNSS constellation are shown in gray in Figure \ref{fig:coverage}. Note that observations from the CYGNSS constellation are limited to approximately $\pm 37^\circ$ latitude due its 35 degree orbit inclination, though upcoming Muon GNSS-R satellites will be launched into polar orbits in order to achieve global coverage.

For the retrieval of SM, the primary GNSS-R observable input to the DL model is the CYGNSS L1a \cite{ruf_cygnss_2018} “power\_analog” DDM, a 17 $\times$ 11 matrix of calibrated reflected power (in Watts). These DDMs represent the measured reflected power originating from scattering regions on the surface surrounding the specular reflection point and recorded in the DDM at specific time delays and Doppler shifts relative to the specular reflection point. Each of the eight individual CYGNSS spacecraft produce simultaneous DDMs from the reflected signals of up to four different Global Positioning System (GPS) transmitters at a 2 Hz rate (previous to July 3, 2019, the CYGNSS DDM rate was 1 Hz), generating a constellation-wide production of order $10^9$ measurements per year.

\subsubsection*{Ancillary Input Data}

Ancillary input dataset selection is guided by the physics of L-band scattering, as well as the characteristics of the transmitted signals and receiver operation, in three categories:

\begin{enumerate}[noitemsep]
    \item Transmitter/receiver characteristics
    \item Surface dielectric characteristics
    \item Surface roughness/topographical characteristics
\end{enumerate}

For the retrieval of surface SM from CYGNSS DDMs, the types of ancillary datasets for the training of a model are listed in Table \ref{tab:inputs}. These datasets are used as direct model inputs and/or as preprocessing filters. Since retrievals of SM from both SMAP L-band microwave brightness temperatures and GNSS-R L-band forward-scattered reflections are governed by closely related emission and scattering physics, respectively, ancillary information used in the retrieval of SMAP SM and packaged in the SMAP L1-L3 Ancillary Static Data (Version 1)\cite{peng} dataset provides a convenient source of most of the ancillary inputs to the DL model. An effort is made to use source datasets without restructuring the file content, though there are often cases where either the source dataset is formatted inefficiently for cloud-based processing or a one-time modification of the dataset would prevent repeating computations. In those situations, the ancillary data is aggregated, modified, and stored in an intermediate state. In situ data (not listed in the table) is strictly used for final validation of retrievals. An additional requirement on the ancillary datasets is that they are available with latency no longer than the CYGNSS data itself so that operational retrieval can be performed as soon as CYGNSS data are available (usually within 48 hours of observation). Therefore, all the ancillary inputs are derived from static or climatological datasets. 

\begin{table}[!htp]\centering
\small
\begin{tabular}{|p{0.2\linewidth}|p{0.16\linewidth}|p{0.08\linewidth}|p{0.4\linewidth}|}
\hline
\rowcolor{cyan}
\textbf{Input Type}&\textbf{Source} &\textbf{Resolution} &\textbf{Information Content}\\
\hline
\textbf{Surface topography}&SMAP Ancillary &3 km &Surface elevation and slope (and their variances) provide large-scale surface roughness and hydrologic context\\
\hline
\textbf{NDVI}&SMAP Ancillary&3 km &Vegetation indices help to detect reflected signal attenuation and scattering by vegetation\\
\hline
 \textbf{Fractional VWC}& SMAP Ancillary& 3 km &Vegetation indices help to detect reflected signal attenuation and scattering by vegetation\\
\hline
\textbf{Surface water fraction} &SMAP Ancillary&3 km &Detection of partial surface water in the FOV that produce strong coherent reflections; used both as an input and/or filter for areas with water fraction greater than 1\%\\
\hline
\textbf{Soil texture} &SMAP Ancillary&3 km &Clay and sand fractions\\
\hline
\textbf{Land cover class} & MODIS(IGBP) & 3 km &Fractional land cover classification map\\
\hline
\textbf{Rx/Tx geometry, reflection location, antenna gain}&CYGNSS L1 Science Data Record v3.2&N/A & Calculation of reflection metrics and context for DDM amplitude/structure \\
\hline
\end{tabular}
\caption{Types of ancillary datasets used in GNSS-R SM retrieval, some of which are used directly as inputs to the model, while others may also play a role in the filtering and preprocessing steps.}
\label{tab:inputs}
\end{table}

Every CYGNSS L1 DDM observation is accompanied by metadata about the reflection geometry and antenna parameters that are critical to interpretation of the measurement. These parameters are used in calibration and the calculation of DDM-derived surface metrics such as reflectivity\cite{chew_description_2020, roberts_deep-learning_2022}. The bistatic geometry and antenna characteristics impact the DDM structure and amplitude and are used as filtering metrics, as well as direct inputs to the model. Latitude and longitude of the specular reflection points are used for match-up purposes but are also included directly as inputs. As discussed in the Feature Importance section, these inputs supplement the ancillary maps of scattering surface characteristics, allowing the DL model to learn regional characteristics of SM. For DL models like those discussed here, including this type of specific geographic location information is only an acceptable practice because the target (SM) is a dynamically varying parameter in most regions. Location information used with fully static (or even slowly varying) target variables will simply train a model to learn the mapping of those variables and therefore should be used with caution.

Surface dielectric conditions influencing the measured DDM include SM, surface water within the spatial extents sampled by the DDM, and vegetation. Surface water, when smooth, produces coherent reflections of L-band signals with DDM features similar to that of very wet soil. As such, the presence of surface water is primarily used as a filter to remove potentially contaminated observations from training and retrieval datasets. A comparison of water masks, including the Global Surface Water Explorer \cite{pekel_high-resolution_2016} and the MODIS-derived water mask provided in the SMAP ancillary data, showed no significant difference in retrieval performance. Therefore, the MODIS-based map is used for simplicity. However, it is important to note that no water mask is able to completely remove all surface water. See Figure 1 in the Supplemental Material for an example of how reflectivity is impacted by the presence of surface water.

Vegetation both scatters and absorbs L-band signals, resulting in attenuated DDMs and with more incoherent-type structure. A climatology of normalized difference vegetation index (NDVI) \cite{peng} is a common, globally available metric for the estimation of vegetation water content. As the water in plants is the primary contributor to scattering from vegetation, a fractional vegetation water content (VWC) input is derived from NDVI using an empirical "stem factor" look-up table (LUT) and a regional land cover type\cite{chan_smap_2013}. A gridded fractional representation of land cover class was derived from the MODIS/Terra+Aqua Land Cover Type Yearly L3 Global 500 m SIN Grid\cite{modis} 2020 dataset and used not only to determine fractional VWC, but also acts as a direct 17-element input to the DL model. Fractional land cover class was derived by summing the contributions of the 500 m pixels for each land cover class within a grid cell and dividing by the total number of pixels for that grid cell, such that each of the 17 land cover classes were assigned a value between 0 and 1 for each grid cell. Due to this fractional representation of land cover, the resulting climatological fractional VWC input is smoothly varying compared to approaches that only use the dominant land cover type. A comparison of fractional VWC and VWC derived only with a single land cover type is shown in Supplemental Material Figure 2. The DL model uses NDVI, VWC, and land cover class all as inputs, as the combination results in higher performance than when each input included in isolation. This is attributed to the fact the VWC is only one of many derivable metrics from NDVI and also has a highly skewed and somewhat regionally discrete distribution compared to NDVI.

Topographic information from the Shuttle Radar Topography Mission (SRTM) digital elevation maps (via the SMAP ancillary datasets \cite{peng}), including terrain elevation, slope, and their standard deviations, are used as ancillary inputs. Surface elevation is additionally used as a filter to remove observations that occur from high-altitude regions where CYGNSS measurements are known to have issues. Small-scale (e.g., centimeter scale) surface roughness, as used in SMAP SM retrievals, is expected to be an important parameter in CYGNSS SM retrievals. But studies of small-scale roughness indicated little impact on model performance, potentially due to parameterization of the roughness dataset studied by the landcover dataset already included as an input. As such, roughness was excluded as an input for this work, but will be revisited in future efforts as newer small-scale roughness datasets are not based on landcover.

As mentioned above, the DL inputs are based on assumed relevance to the specific case of L-band surface scattering physics. That said, physical relevance alone is not sufficient to guarantee impact on model training. If a physically-relevant parameter is not statistically significant or the variability of that particular input is low, there will be minor impact on the training process. Methods to handle these situations include altering how the model learns by introducing biases to underrepresented populations. These challenges and effects are discussed further in the Feature Importance subsection, which details how the final set of DL ancillary inputs was determined.

\subsubsection*{Determining Grid Cell Size of Ancillary Inputs}

Ancillary land surface parameters used as inputs to the DL model are available at a variety of spatial resolutions and gridding schemes. It is important to match, to the greatest extent possible, the spatial resolution of the ancillary data to that of the GNSS surface reflected signal. However, the spatial resolution of GNSS-R observations varies depending on the roughness of the reflecting surface, with smoother surfaces producing reflections with smaller spatial footprints, and vice versa. For areas with RMS surface heights < 5 cm, the smallest theoretical spatial footprint for a GNSS reflection recorded by a LEO receiver is ~0.2 – 1 km in diameter, depending on the incidence angle. Because CYGNSS integrates each observation along its track for 0.5 sec, this footprint is smeared across the surface such that the smallest possible footprint is $\sim$0.5 x 3.5 km. For rougher reflecting surfaces, spatial footprints are much larger—on the order of 20 – 25 km \cite{clarizia_wind_2016}.

\begin{figure}[H]
    \centering
    \includegraphics[width=1\linewidth]{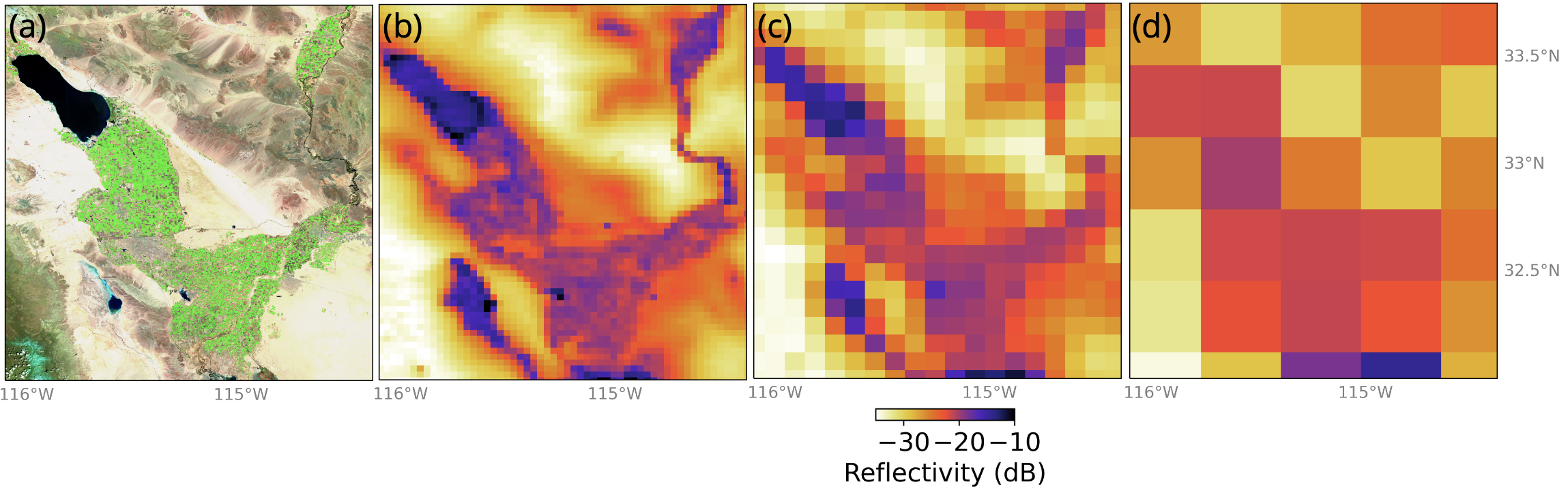}
    \caption{False-color image from Sentinel-2 over the Salton Sea region in southern California (a) and long-term averages of CYGNSS reflectivity gridded to 3 km (b), 9 km (c), and 36 km (d).}
    \label{fig:GriddingFigure}
\end{figure}

Unfortunately, the small-scale roughness of the land surface is difficult to measure, as this centimeter-scale roughness is determined by processes like animal burrowing, rain surface rilling, agricultural tilling, etc. and is thus not captured by available global digital elevation models. Some direct measurements exist\cite{Hornbuckle, campbell_intercomparison_2022} and indicate RMS surface heights on the order of 2-3 cm for typical rangelands and agricultural areas. Given the large span of possible spatial footprints depending on surface roughness and the dearth of direct small-scale roughness measurements, many in the GNSS-R community have developed methods of quantifying whether a reflection is predominantly coherent (i.e., the reflecting surface is smooth) or incoherent (i.e., the reflecting surface is rough). However, results from these studies have also varied widely, with some studies determining that most reflections coming from the land surface are incoherent\cite{Al-Khaldi} and others determining that most reflections from the land surface are coherent\cite{dong_evaluation_2021}.

Here, the spatial resolution of the GNSS surface reflection is defined as the grid cell size at which further coarsening would result in a loss of spatial information. Figure \ref{fig:GriddingFigure} shows one example of CYGNSS reflectivity observations over the region southeast of the Salton Sea gridded at 3 km, 9 km, and 36 km (representing the most commonly used EASE-Grid 2.0 gridding schemes\cite{brodzik_ease-grid_2012}). Comparing the gridded reflectivity maps to a high-resolution optical Sentinel-2 image in Figure \ref{fig:GriddingFigure} a), the delineation between desert, agricultural, and urban areas in the 3 km reflectivity images is clear, though much of this information is lost when gridding to 9 km (all detail is lost when gridding to 36 km). Thus, ancillary datasets at a 3 km scale are used to match with each CYGNSS observation as inputs into the model.

\subsubsection*{SMAP Data Filtering}
SMAP quality flag filtering practices are identical to those detailed by Roberts at al.\cite{roberts_deep-learning_2022}, where some flagged SMAP data are retained but clearly erroneous SMAP data are selectively removed (e.g., retrievals flagged as "unsuccessful"). This allows for training over the majority of the land surfaces where SMAP data are available, while avoiding the propagation of significant issues in the SMAP retrievals. As an example, SMAP retrievals over the majority of South America and Central Africa are flagged as “not recommended”, but we choose to still use these retrievals to avoid having regions totally devoid of SMAP data. A likely outcome of using this data is an upper bound to model performance on par with SMAP's low performance at these locations. As such, these locations are flagged accordingly in the resulting gridded products. Alternative solutions to using these lower quality training targets are being investigated, which will likely be included in a future version of this data product. Additionally, SMAP surface flags indicating the presence of significant precipitation are also used as a filter to avoid training on measurements observed during rain events.

\subsection*{Dataset Ingest and Organizational Logic}

The datasets described in the previous sections are ingested from external sources and reside in a "data lake". These data are arranged and, in certain cases, lightly preprocessed to make them efficient for cloud-based processing. The next step in the processing pipeline is creation of a data warehouse of "analysis-ready data", structured specifically for retrieving SM from GNSS-R observations. Here we describe the aggregation and generation of both databases.

The data ingestion pipeline (green in Figure \ref{fig:retrieval_flow}) has been designed around the concept of matching contextual and target information to each GNSS-R measurement. For CYGNSS data, the steps in the process are:

\begin{enumerate}[noitemsep]
    \item Aggregate source datasets into the data lake. Temporally varying datasets (including CYGNSS L1 data) are operationally pulled daily.
    \item For each GNSS-R DDM measurement in a CYGNSS L1 data file (which are distributed as daily files, one each per CYGNSS satellite), create a “sample” object containing all provided information per DDM (e.g., reflection time, location, quality flags, etc.).
    \item For each of these samples, match the reflection time and location to corresponding surface information from available ancillary datasets, including potential target datasets. Performing steps 2 and 3 for all CYGNSS L1 data creates the data warehouse of analysis-ready daily sample sets.
    \item A composite dataset for training is compiled by filtering according to the nature of the desired retrieval (e.g. SM occurs over land)
\end{enumerate}

Figure \ref{fig:retrieval_flow} highlights steps for the SM retrieval process (marked with black numbers). For steps 2 and 3, values are pulled from the data lake containing all of the sourced L1 CYGNSS files and relevant ancillary datasets. These are, when possible, left in their original file format. Potential modifications include standardizing the data gridding  across all datasets into either EASE-2.0 grids or degree-regular grids of various resolutions. Step 3 is made efficient by leveraging this standardized gridding and determining the indices of each grid type during sample generation. Finding matchups is simply performed by indexing into a two-dimensional array (three-dimensional array if temporally varying). These steps are effectively a remapping of the source data into an analysis-ready format. This data warehouse can be queried for data fulfilling a number of requirements, making the generation of a training dataset trivial. For CYGNSS, the data warehouse is organized by day and per satellite due to the nature of the original CYGNSS L1 data, but data from Muon GNSS-R satellites will use a shorter time window to achieve lower latency processing.

In step 4, the filtering of the samples is specific to the geophysical variable being retrieved (e.g., SM must be retrieved over land), but filtering is also a tunable metric for model performance. By excluding excessively challenging or problematic data, model performance will likely improve on the residual target space. This is, in part, a quality control step, but it also allows for tuning models specific to a region of interest or possessing skill related to specific surface characteristics (e.g., a model tuned to SM retrieval over a specific region or land cover type).

\subsection*{Model Development}

Major elements of model development include the input and target dataset down-selection and preprocessing, model architectural choices, training, and performance analysis (i.e., the blue components of Figure \ref{fig:retrieval_flow}). The previous section provided details on the underlying samples that populate the GNSS-R data warehouse, which act as the starting point for model building. Here, we detail additional sample set filtering based on considerations relevant to SM retrieval and then how standard machine learning data preprocessing steps are applied to these samples prior to model training. We then provide details on the DL model architecture and explain the technical details of how the model is trained. Following this is a discussion on how the model is evaluated against the target, which informs subsequent changes made to the model architecture and/or preprocessing steps before selecting a final model. The SM retrievals are evaluated against in situ ground stations measurements of SM, covered in the Technical Validation Section. Finally, the quantification of input importance, model sensitivity to input variation, and model training uncertainty are presented.

\subsubsection*{Dataset Filtering, Splitting, and Preprocessing}

CYGNSS L1 data acts as the starting point for building a training dataset, and filtering these data is the first step. Version 3.2 of the CYGNSS L1 data includes two sets of quality flags to indicate potentially erroneous measurements. Certain flags are used to remove data of no interest to retrieval of SM, such as measurements made over the ocean or during instrument calibration. Table 1 in the Supplemental Material lists the CYGNSS quality flags used and the percentage of the CYGNSS observations removed via each flag, as well as the additional filtering listed below. Several internally developed filters based on the CYGNSS data include a threshold on the DDM SNR, (ddm\_snr > 1 dB), a requirement that the specular point antenna gain of the receiver is greater than 1 dB, and a maximum reflection incidence angle of $65^\circ$. The RFI flag included with the CYGNSS data was developed for use over the ocean and fails over higher elevation land surfaces, so a replacement RFI flag was developed that detects RFI when entire Doppler columns of the DDM are five times higher in amplitude than the DDM noise floor.

While target data is not needed when performing retrievals, a fundamental requirement for building a training dataset is that there are coincident CYGNSS and SMAP measurements. As mentioned in the previous section, in order for a CYGNSS observation to be used in training, a colocated and same-day SMAP retrieval must also exist. Here, colocated means the CYGNSS specular reflection point location fell within a populated cell of the 9 km SMAP AM grids for a given day.  We found that narrowing the time window limits from $\pm12$ hours to $\pm4$ hours did not result in model improvements worth the resulting reduction in sample count. In addition to colocated SMAP measurements, filtering criteria required valid values of all of the ancillary inputs in Table \ref{tab:inputs}. A threshold of less than 1\% surface water fraction within the DDM extents was implemented to avoid training data contaminated by known bodies of water that result in highly coherent reflections. Additionally, a maximum surface elevation of 3000 m was used to remove DDMs potentially truncated by CYGNSS's elevation-constrained onboard digital elevation map (DEM). 

Using the above criteria, a sample set was constructed from the CYGNSS v3.2 L1 data record from August 2018 through March 2024. Note that this time window includes two changes that could theoretically affect the CYGNSS observations: the switch from 1 Hz to 2 Hz sampling on July 3, 2019, and the activation of Flex Power mode IV on the GPS constellation\cite{steigenberger_flex_2019}. Both of these changes should be compensated in the L0/L1 processing\cite{cygnss_L1}, but we mention it here for completeness. With the above filtering, we constructed training, development, and validation datasets using the following windows:

\begin{itemize}[noitemsep]
    \item Training window: January 2021 through December 2022:~ 6.6e7 samples
    \item Development window: January 2023 through December 2023: 2.1e7 samples
    \item Validation window: August 2018 through December 2020: 6.6e7 samples
\end{itemize}

Note, the naming convention for the sample splits differs slightly from what is more common in machine learning literature, where sample splits are referred to as “train,” “validation,” and “test.” We refer to our unseen holdout set as “validation” rather than “test” to better align with nomenclature used for geophysical model development. We have found that randomly-sampled splits result in models that are highly over-fit to the training data due to the nature of GNSS-R data where adjacent samples are highly correlated. Therefore, temporal separation of the training, development, and validation sets has been used for all GNSS-R model training. Note that all windows are intentionally set to at least a full year to ensure that seasonality is observable, and the splitting is chosen for optimal overlap with key in situ validation datasets only available during certain windows of time. Finally, the data available for training and development has been downsampled by a factor of two to limit the memory requirements for model development using the above windows. This is implemented in a manner that effectively reduces the CYGNSS sample rate, achieving a similar distribution of training samples across the surface of the Earth but slightly more separated along-track.

The preprocessing of data prior to model training ensures numerical stability during the learning process. Avoiding situations where one input is many orders of magnitude larger than another allows for the training process to more easily find global minima. As such, the inputs are normalized from 0-1, or standardized from -1 to 1, depending on the input distribution. It was also observed that the DDM amplitude distribution varied slightly between the CYGNSS spacecraft; therefore, DDMs were standardized using means/variances as calculated for each spacecraft. Note that normalization is calculated after quality controlling the datasets.

\subsubsection*{Network Architecture}

Within the retrieval framework described in the previous section, the specifics of the model architecture act as a modular component, both at a model-type level and at an input-selection level. We have explored several model paradigms ranging from XGBoost to fully dense neural networks, but we have primarily focused on using convolutional neural networks (CNN) as a mechanism to ingest the full DDM array due to the inherent ability of CNNs to extract structural information from arrayed data. Different GNSS-R reflection conditions can drive a number of structural features in DDMs, which the convolutional layers of the network learn to extract and then map to weights at the output dense layer. These weights are then concatenated with ancillary contextual information about the particular surface characteristics and receiver configurations. A diagram of the network topology is shown in Figure 3 of the Supplemental Material.

Our early models used an intentionally simple convolutional architecture to focus on deriving performance mainly from the choices of input data and processing, but we explored more advanced network architectures that have the potential to establish more complex relationships and achieve better results. The current model includes two residual blocks, and each residual block includes two convolutional layers with 3x3 filters, leaky ReLU activation functions, and a skip connection. The skip connection allows the outputs of earlier layers to skip past subsequent layers and be added to the output of later layers. A main benefit of the skip connections is that it opens up an alternative flow of information that promotes the reuse and preservation of earlier features in later operations, which may be important in the context of learning for a relatively simplistic input array, the DDM, when compared to traditional, more complex image tasks. The leaky ReLU activation helps prevent the occurrence of "dying ReLU," where neurons with large negative values end up evaluating to zero following ReLU and resulting in a zero gradient that stops weights from updating. Leaky ReLU introduces a small, tunable negative gradient that helps alleviate this problem. 

Max pooling is applied to the output from the convolutional layers, wherein the pooled features are flattened and concatenated with ancillary features. This architecture passes the ancillary inputs through a single, dense layer prior to concatenating with the convolutional outputs, similar to the architecture described in Asgarimehr et al.\cite{asgarimehr_gnss_2022}. This step performs a type of learned dense featurization specific to the ancillary inputs, much like the residual blocks perform learned featurization to the DDMs. The combined flattened convolutional outputs and dense ancillary layer outputs are passed through two dense layers, each with a small amount of dropout (less than 5\% probability of being dropped) to mitigate overfitting, before being passed to the final output layer. The output layer is a single node that uses a softmax function to restrict the output, SM, from having a non-physical negative value. Each of these changes from the early, simplistic network was tested in isolation to ensure it increases the model's capacity to capture SM dynamics without overfitting or driving unnecessary computation.

\subsubsection*{Model Training}

Model training is performed using the standard approach of dividing the dataset into training, development, and validation subsets (described above). Here, training data are fed to the model for “learning.” The development set is used during the training process to benchmark the performance improvement and detect overfitting. The validation set is used at the end of this process to determine how well the model performs on totally new, unseen data. 

Model training is performed using the TensorFlow/Keras libraries. In general, models are trained over 30 epochs when testing changes to the inputs or architecture, while the final model was trained for 65 epochs and did not exhibit signs of overfitting. The loss function used is the mean squared error between the target and prediction. Optimal ranges for the learning rate were determined using an optimal learning rate routine\cite{smith2017cyclicallearningratestraining}. This approach yielded a range of 1e-5 to 1e-3 that stably produces sufficient loss in acceptable time.

\subsubsection*{Performance Evaluation}

The performance (i.e., skill) of a given model is studied using several methods in subsequent stages, looking at the performance of the predictions against two different sources of "truth": target data from the validation window and in situ data from ground stations within the International Soil Moisture Network (ISMN)\cite{dorigo_international_2021} and SMAP Core Validation Sites (CVS)\cite{COLLIANDER2017215}. Performance against the validation set is the baseline metric used to understand how well the models are learning.

Here we discuss the validation data as it pertains to model training evaluation and save the discussion of in situ data analysis for the Technical Validation section. More in-depth comparisons to SMAP are also included in the Technical Validation section. Here we also compare our Muon SM product to the official v1.0 CYGNSS SM product\cite{ucar} that uses the University Corporation for Atmospheric Research/University of Colorado algorithm (hereafter abbreviated as "UCAR"). There are several differences between our algorithm and the UCAR algorithm, though the two primary differences are that, at the time of writing, the UCAR algorithm uses v2.1 of the CYGNSS data (we use v3.2) and it relies on linear regressions between CYGNSS and SMAP matchups on a per grid cell basis that outputs a 36 km gridded product. Note that after the preparation of this paper was completed, a newer version of the official CYGNSS SM product was released, which uses CYGNSS v3.2 data as well as the 9 km SMAP SM retrievals in their regression analysis, and future work will include an assessment of our performance against the updated version of the official CYGNSS SM product.

The validation samples are CYGNSS GNSS-R observations colocated with the target variable that are intentionally not used in training the model. Explicitly, they are from a separated window of time than the training or development sets that were used in the training process. While the training and development test splits were downsampled to limit resource utilization, the validation window is evaluated in full. They provide an unbiased estimate of the model’s ability to predict SM, as estimated by SMAP, on new and unseen GNSS-R data. Note that this is fundamentally a measure of how well the model was trained to reproduce SMAP SM estimates, but it is not necessarily a measure of how well the model outputs true soil moisture. Any errors in the SMAP target dataset are not measurable with this analysis.

\begin{figure}[H]
    \centering
    \includegraphics[width=1\linewidth]{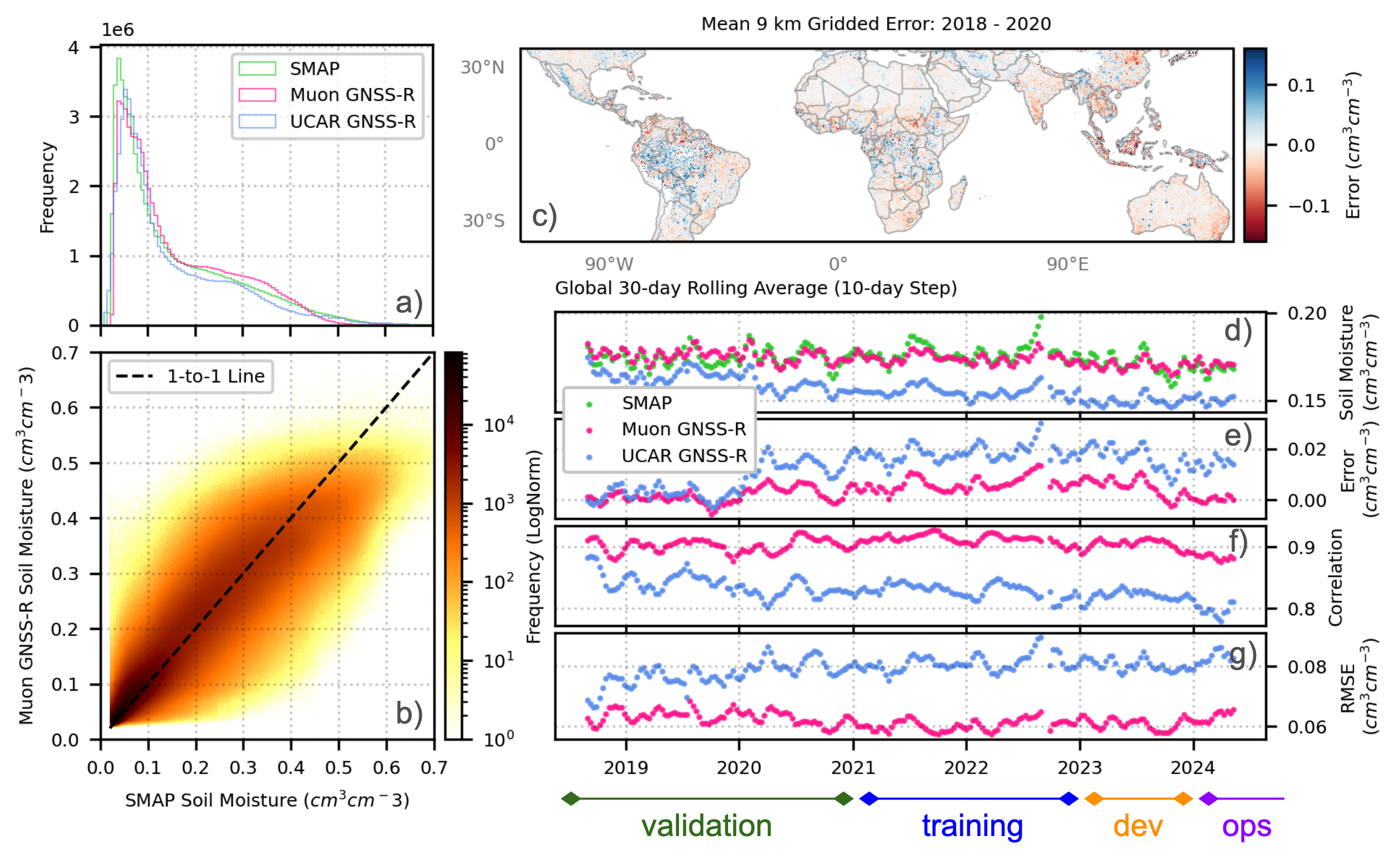}
    \caption{Evaluation of the model performance against the target and comparison to the UCAR product. a) distributions for the target SM (green), predicted SM (pink), and UCAR SM (blue). b) the relationship between predictions and targets with color on a logarithmic scale. c) the global difference between prediction and target averaged over the validation window. d-g) rolling average windows of SM from SMAP (green), the Muon product (pink), and the UCAR product (blue), error from SMAP, correlation with SMAP,and RSME from SMAP, respectively. The colored lines under the time series plots indicate the different data windows.}
    \label{fig:performance_eval}
\end{figure}

Figure \ref{fig:performance_eval} a) compares the CYGNSS SM retrieval distribution (pink) to the target SMAP SM distribution (green) and the UCAR product (blue). Figure \ref{fig:performance_eval} b) shows the correlation between retrieved CYGNSS SM and target SMAP SM, yielding a Pearson correlation coefficient of 0.907, a slope of 0.84, and an RMSE of 0.053. One aspect to note is the disagreement at higher SM values (greater than 0.45 $m^3/m^3$). This is likely due, in part, to the under-representation of this population of the distribution in model training. We have explored methods such as sample weighting and distribution flattening in an attempt to create better agreement for these higher SM values, but as the SMAP values are likely overestimates driven by surface water contamination, we applied no such adjustment and expect disagreement in this range. This is an example of where we actively choose to deviate from our target values based on what we believe to be potential errors in that dataset. Figure \ref{fig:performance_eval} c) shows a global average over the validation window (August 2018-2020) of the difference between the target and predictions values. SMAP retrievals are generally wetter in the Amazon and the Congo, but they are drier in parts of the Sahel, India, China, and Indonesia. The drier bias of the model's retrievals are examples of this divergent behavior.

Figure \ref{fig:performance_eval} d-g) shows results of an analysis of the full data record time series using a 30-day rolling average with a 10-day step. Figure \ref{fig:performance_eval} d) compares the SM retrievals from SMAP (green), Muon's CNN model (pink), and the UCAR product (blue), while e) shows the differences from SMAP. While both Muon and UCAR SM products capture the general trends, the Muon SM shows a reduction in bias from SMAP and better correlation with temporal variations. The deviation seen in the UCAR SM product after Feburary 2020 is due to the activation of the Flex Power mode IV on the GPS satellites, and the UCAR SM product likely improves in the newer version of the algorithm. Figures \ref{fig:performance_eval} f,g) show the correlation and RSME (respectively) of each GNSS-R SM product against SMAP, indicating consistent improvement in both statistics versus the official v1.0 CYGNSS SM product. An important final aspect of all four plots is that they span the training, development, validation, and operation windows, and the Muon SM product shows similar characteristics across all four windows, indicating the model generalizes well even outside of the training window.

While perfect agreement between the model predicted GNSS-R SM and the target SMAP SM is technically the goal of the training process, problematic features of the target SMAP SM can also end up being represented by the model. This is partially limited by selective filtering of SMAP data, but inherently perfect agreement with SMAP SM is probably not the correct goal. As an example, we explored inputting an annual average of SMAP SM values as a "prior" for each retrieval location  to allow the model to learn from an expectation of SM variations measured by SMAP. This markedly improved our performance against SMAP during validation but also resulted in spatial trends that were very "SMAP-like." Some of these trends showed the impact around large bodies of water of SMAP's much larger 36 km sensing footprint  (resulting in generally wetter GNSS-R SM retrievals), as well as a general smoothing of small-scale spatial SM features. We therefore opted to exclude this bias input, which reduced our overall performance against SMAP, but avoided potentially unrealistic biases. As previously mentioned, determining the best way to filter and ingest SMAP data is an ongoing area of research.

\subsubsection*{Input Importance and Sensitivity Studies}

As noted earlier, two factors can affect how important a given input is to model performance: 1) the physical relevance of the input to the retrieval being made; and 2) the statistical significance of the input's variation. The former is whether the input provides important context to understand the reflection conditions, especially if it can help disambiguate particular aspects of the observation that could be attributable to the desired output or the input in question. The latter is whether this particular input demonstrates substantial variation a sufficient number of times to allow the model to learn. For example, if a certain input is highly impactful to SM but rarely varied (e.g., the presence of an extensive wildfire), this input would likely have little importance in the retrieval process. Reciprocally, if an input varied frequently, but was totally unrelated to SM, this would also have little importance. An input must have both of these properties to be useful to the model.

\begin{figure}[H]
    \centering
    \includegraphics[width=1\linewidth]{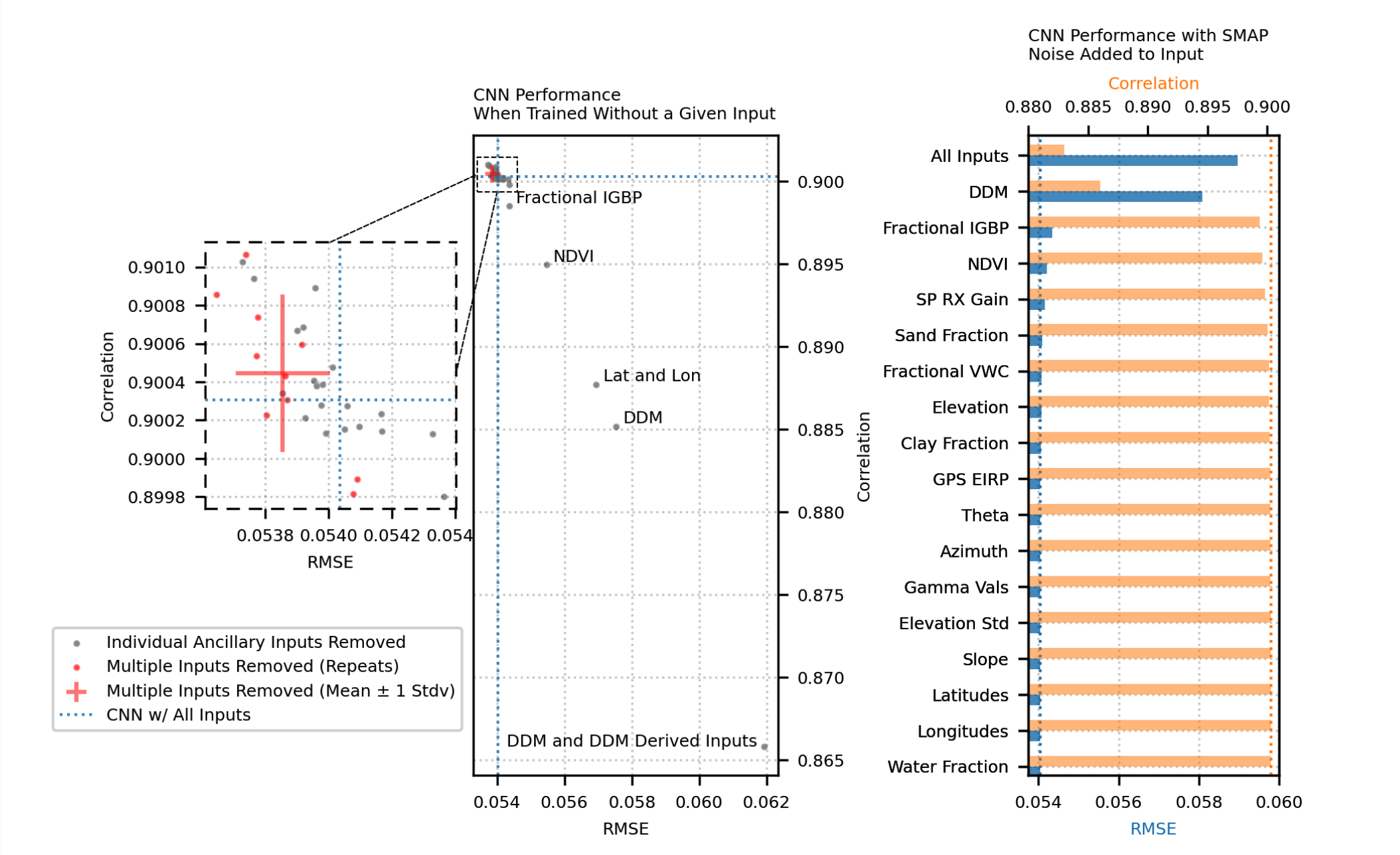}
    \caption{Results of input-ablation study (left/middle), and input sensitivity study (right). The bar chart lists the sensitivity of all the inputs used in the final model.}
    \label{fig:feature_importance}
\end{figure}

Several methods were explored for understanding the relative importance of the inputs to the models, and input ablation studies were found to be the best way to fully remove an input's contributions to a model without observing distribution dependencies to which alternative methods, such as input permutation (shuffling the values of a given input), are sensitive. Input ablation is a process where multiple models are trained, each missing one input. A list of all the inputs used by the final model is given by the bar chart labels of Figure \ref{fig:feature_importance}. Inputs that are of low importance will have little impact on the performance of the model as measured by the validation set, while inputs of high importance will cause a degradation in performance when removed. Here "performance" is measured as correlation and RSME against the target variable. This process requires the training of many models, so it was performed using a reduced dataset (1.7e7 samples) compared to the training dataset used for the final model.

A baseline for performance was set by a model trained on the complete set of available inputs. A separate model was trained for each of the inputs, excluding (ablating) them one by one. The performance of the baseline CNN is indicated by the dashed blue lines on the Figure \ref{fig:feature_importance} (left), while each grey marker corresponds to the performance of an individual ablation model. In a few cases, multiple inputs were collectively excluded during an ablation run given their natural relationship to one another, e.g., latitude and longitude and the DDM and DDM derived inputs. The most noticeable decrease in performance occurred when the DDM and all DDM derived inputs were removed from the CNN. This particular ablation test helped demonstrate the performance achievable by learning global SM biases and the relevance of inputs varying in time and space to improve upon those biases. Another interesting case is where the DDM is removed, but the DDM-derived metrics are retained (point marked "DDM" in Figure \ref{fig:feature_importance} (left)). The performance gap between this point and the point with all DDM information removed quantifies the benefits of including structural information from the DDM array instead of simply the peak value. The other label markers that indicate important inputs include location Lat/Lon pairs, NDVI, and land cover (IGBP). The model is likely learning regional biases from the location and land cover information and seasonal trends from the climatological NDVI values. 

The majority of models with ablated inputs cluster together near the performance of the baseline CNN, showing that the relative importance of each of these inputs is low when removed individually. However, removing groups of these "low value" inputs together results in a noticeable performance decrease. This behavior is likely due in part to multicollinearity associated with the inputs, i.e., the model is able to adjust its weights to make up for the lack of one input by using the remaining correlated inputs, and is a drawback to the approach of this ablation study. The red dots in the upper left inset of Figure \ref{fig:feature_importance} show the performance of ten repeated runs of a particular model (discussed more in the following subsection). The red cross indicates the mean and standard deviation of these runs and gives insight into the variance in performance of each ablation model.

\subsubsection*{Model Uncertainty Quantification}

Model uncertainty can be broken into two categories: 1) how the uncertainties in inputs affect the model output; and 2) how the model training process itself involves some variance. To understand the extent to which uncertainty in inputs influences the retrievals, we started with the same baseline model as described in the ablation study. Using this trained model, random noise was added to each input, and predictions were made. This process was carried out individually for each input. For inputs with known uncertainties, those uncertainty distributions were sampled directly when determining the quantity of noise to add to the input for each sample. In the case that variances were not quantified, noise was sampled from a normal distribution with a range of $\pm$5 percent at a 90 percent confidence level, where the sampled value was applied to the given observation’s input to correctly scale the noise value. For the DDM case, this sampled value was the DDM peak, and noisy values were applied to the entire 17 x 11 array. The bar chart of Figure \ref{fig:feature_importance} (right) shows the correlation and RMSE for retrievals when noise is individually added to each input that is used in the final model and compared to the baseline performance without noise, represented by the vertical blue and orange dotted lines. The top category in the bar chart in Figure \ref{fig:feature_importance} indicates the change in performance when predictions were made when all inputs have added noise. This is the most conservative case in terms of the performance from the model assuming all inputs have some quantity of added noise, and given the specified noise values, resulted in a degradation in correlation of less than 0.02 and an increase in RSME of less than 0.006 cm$^3$ cm$^{-3}$, which is small. The drop in correlation and increase in RMSE are reasonable given the added noise levels and indicate a level of stability to input variance. A nuisance of this test is that many of these inputs were static maps, so this noise is not representative of what the model will actually see in practice, but more so of possible departures from accurate representation.


To quantify how the models can vary during training, we performed an ensemble training process where 10 models were trained using identical inputs and architecture, but the training process was initialized with a different random seed. This test quantifies the uncertainty in the SM retrievals due to model variability resulting from stochasticity during training. To address this question (as well as strengthen the interpretability of the ablation study), a model with "best" inputs was selected and trained 10 times using different random initialization seeds. Due to random aspects in the training process, weight initialization, and batch sampling within epochs, training the same neural network multiple times with a different random seed will result in slight differences between the networks’ learned weights. These differences should not amount to fundamentally large differences in the predictions. Across the repeat training runs, we see a tight distribution of the CNNs’ RMSE and correlation. The center of the red cross (Figure \ref{fig:feature_importance}) corresponds to the mean RMSE and correlation of the 10 identical models, while the length of the vertical and horizontal lines corresponds to ±1 standard deviation of the correlation and RMSE, respectively. When considering the variability between the shot-to-shot predictions made by the 10 models, the mean standard deviation between the 10 predictions was 0.008 cm3/cm3, with a standard deviation of 0.005 cm3/cm3. With this relatively small level of variance between models, we trained a final model for 65 epochs which was used for the operation model in the pipeline (dark purple, Figure \ref{fig:retrieval_flow}).

\subsection*{Level 2 Product Generation}

Following the training process (described in previous subsection), and the validation analysis (described in the Technical Validation section), the trained model is used to make retrievals on the full time series of CYGNSS L1 v3.2 data. Here, sample sets for each day and spacecraft are pulled from the data warehouse and fed to the model for inference. The retrievals generated from the model, as well as all of the ancillary data from these sample sets, are propagated through to the prediction dataset (with the exception of the DDM arrays to reduce storage volume). These outputs, which form the Level 2 products, are subsequently broken into individual track-wise (per CYGNSS spacecraft and GPS transmitter) and aggregate daily track-wise files for each spacecraft. 

CYGNSS quality flags are propagated through to the retrieval process, and are included in the L2 product. Note that all of the spacecraft and land surface ancillary data is provided with the retrievals for users to better understand the context in which the retrieval was made. This is an effort to enable users to gain insight into how retrievals could be related to valid or invalid assumptions about the reflection conditions and allow users to apply  their own filtering and quality control. Therefore, a user could generate their own Level 3 retrievals adhering to criteria differing from what we used to generate the Level 3 product described in the next section.

\subsection*{Level 3 Product Generation}

The L2 product described above is generated on a per-track basis, and the individual SM estimates available in the tracks are needed by applications such as data assimilation. But many users are interested in gridded datasets with integrated/averaged measurements from some specified duration of time. Because the model was trained on 9-km gridded SMAP SM retrievals, we also grid our L3 SM product to the 9-km EASE-2.0 grid at hourly and daily intervals.

During each hourly time interval, there tends to be between 0 - 5 samples within a particular 9 km grid cell. We investigated three different ways of aggregating the individual retrievals to create the gridded product: equal weighting, nearest neighbor, and inverse distance weighting. There was little to no quantitative difference in retrievals generated with any of the three approaches, likely due to the small number of observations being averaged within a single grid cell. Ultimately, we chose to use the equal weighting method, as it slightly outperformed the other two gridding methods when comparing the gridded retrievals against the sparse in situ sites. Using this gridding, retrievals from all of the individual spacecraft are combined into separate hourly and daily L3 products.

As SMAP SM retrievals were used to train the model, quality flags were developed with the aid of SMAP ancillary gridded data\cite{peng} for the L3 data. With the exception of the dense vegetation flag, which is based on a climatology, all other flags are static. These flags are preliminary and may change with future release versions. Table \ref{tab:surface_flags} lists the current surface flags.

\begin{table}[!htp]\centering
\small
\begin{tabular}{|M{0.05\linewidth}|M{0.15\linewidth}|M{0.1\linewidth}|p{0.6\linewidth}|}
\hline
\rowcolor{cyan}
\textbf{Bit} &\textbf{Surface Condition} &\textbf{Threshold (T1)} &\textbf{Interpretation} \\
\hline
0 &Coastal proximity &10 km &SM retrieved regardless of proximity to coast, but flag is set to 1 (uncertain quality) when coastal proximity < T1. \\
\hline
1 &Urban fraction &0.25 &SM retrieved regardless of urban fraction, but flag is set to 1 (uncertain quality) when urban fraction > T1. \\
\hline
2 &Permanent snow/ice &15 &Flag is set to 1 when dominant IGBP class = T1. \\
\hline
3 &Elevation &3000 m &Flag is set to 1 and SM is not retrieved when elevation > T1. \\
\hline
4 &Dense vegetation &5 kg m-2 &SM retrieved regardless of VWC, but flag is set to 1 (uncertain quality) when VWC > T1. VWC is based on a climatology of NDVI. \\
\hline
5 &Frozen ground &- &(currently empty) \\
\hline
6-7 &Reserved &- &(reserved for future flags) \\
\hline
\end{tabular}
\caption{SMAP-derived quality flags (surface\_flag field) for the L3 product}
\label{tab:surface_flags}
\end{table}

\section*{Data Records}

In this section we discuss how the data records will be made available for Version 1.0 of the Muon Space GNSS-R Surface Soil Moisture Product, along with expected latencies, data formats, and data file organization. Future minor revisions of this product which retain all existing fields will increment the decimal portion of the version number, while more major updates will increment the integer portion of the version number. 

We start with a brief review of the daily process for generating GNSS-R SM retrievals from CYGNSS data:

\begin{enumerate}
    \item The PODAAC server is queried for new L1 files which are downloaded into the Muon data lake
    \item The L1 files are converted to filtered and preprocessed sample sets to prepare for interfacing to the machine learning model 
    \item Predictions of SM are made using the trained model and also stored in the Muon data lake
    \item L2 and L3 product files are generated from the prediction data and pushed to an accessible data repository
\end{enumerate}

CYGNSS L1 data is usually available with a 2-3 day latency, which is the most significant latency in generating the SM product. For Version 1.0, all of the ancillary information comes from either static or climatological datasets that add no latency. The above process runs daily and is repeated using a sliding window in time to ensure that partially available or delayed datasets are backfilled. Additionally, if data that is not required for retrievals (such as SMAP) becomes available during this window, these fields are also updated in the the L2 files.

\subsection*{Data Access}


Access to the CYGNSS-only version of the Muon Space GNSS-R Surface Soil Moisture Product will be made available for public download via two mechanisms. The majority of these data (as of publication) is hosted on Zenodo, which includes the windows discussed in this paper up until September 26, 2024. This is accessible via the Zenodo data record page \url{https://zenodo.org/uploads/14172456}. This repository will allow users to explore the data entirely for free. An operational version of this dataset (including dates past September 26, 2024) currently exists in a Muon hosted data repository, and will be linked to from the above Zenodo page, but will potentially require an AWS S3 account and may incur AWS costs associated with data access depending on how the data is downloaded. Efforts are underway to make this operational dataset fully free/accessible, but were not finalized at the time of publication. Check the Zenodo page for updates on how the entire dataset can be accessed. Additionally, the Zenodo page includes a link to a git repository with code demonstrating how to read and visualize the data files. The below subsections detail the structure of the data contained in these repositories.

\subsection*{GNSS-R L2 CYGNSS Surface Soil Moisture Product Format}

The L2 SM product is provided as ungridded, trackwise retrievals in netCDF4 format, with an assumed spatial footprint of approximately 3-9 km depending on the surface roughness and representing the average SM for the top 5 cm of the soil column (i.e., near-surface SM). L2 files are produced for each day of operations; one daily file containing all the aggregated trackwise retrievals for a given satellite on a given date. Files are produced for each calendar day the satellite is in operation.

\noindent The directory/prefix structure for SM retrievals is as follows:

\path{/{version}/trackwiseSoilMoisture/{satelliteID}/{YYYY-MM-DD}/}

The "version" (currently "v1.0") is the released product version. The satelliteID is a two letter mission identifier followed by a three-digit number identifying the spacecraft. In the case of CYGNSS data, the identifiers are: \path{CYxxx}, where \path{xxx} can be anywhere from \path{001-008}. The final subdirectory identifies the date in year-month-day format.

\noindent File name format for the L2 files is given as:

\path{aggregateSoilMoisture_muon_{satelliteID}_{YYYYMMDD}_{version}.nc4}

Missing values are all indicated by -9999. A complete table of the product contents is available in the Supplemental Material in Table 2.

\subsection*{GNSS-R L3 CYGNSS Surface Soil Moisture Product Format}

The L3 SM product contains daily and hourly gridded averages of the L2 CYGNSS SM retrievals provided on a 9 km EASE-2.0 grid and stored in netCDF4 format. Files are produced for each calendar day that data is available from at least one CYGNSS satellite. Table \ref{tab:level_3_file_contents} gives a summary of the variables contained within the retrieval files. Missing values are all indicated by -9999.

\begin{table}[!htp]\centering
\small
\begin{tabular}{|M{0.2\linewidth}|M{0.1\linewidth}|p{0.6\linewidth}|}
\hline
\rowcolor{cyan}
\textbf{Variable} &\textbf{Units} &\textbf{Description} \\
\hline
latitude & degree north &Latitude of the grid cell centers \\
\hline
longitude & degree east &Longitude of the grid cell centers \\
\hline
soil\_moisture\_level3 &m3 m-3 &Soil moisture average per cell \\
\hline
surface\_flag &unitless &Bit flag that records ambient surface conditions for the observation. \\
\hline
\end{tabular}
\caption{Contents of L3 product files.}
\label{tab:level_3_file_contents}
\end{table}

Although the files contain global grids , due to the latitudinal constraints of CYGNSS, nearly all SM retrievals fall within +/- 38 degrees latitude. All times are in UTC. Each hourly file contains averages of observations that were in the interval [$time\_start$, $time\_end$), while each daily file contain averages of all observations within that calendar day. Global attributes $time\_start$ and $time\_end$ are contained within each file.

\noindent The directory/prefix structure for the hourly and daily SM retrievals is as follows:

\path{/{version}/griddedSoilMoisture/{filetype}/CYGNSS/{YYYY-MM-DD}/}

The file type can be "hourlySoilMoisture" or "dailySoilMoisture".

The "version" (currently "v1.0") is the released product version. The final subdirectory identifies the date in year-month-day format.

\noindent File name format for the hourly and daily SM files:

\path{hourlySoilMoisture_muon_CYGNSS_{YYYYMMDDTHH}Z_v1.0.nc4}

\path{dailySoilMoisture_muon_CYGNSS_{YYYYMMDD}_v1.0.nc4}

In the hourly case, the time field YYYYMMDDTHHZ refers to the beginning of the hourly window for which observations within a grid cell were averaged.

\section*{Technical Validation}

Here, we detail the retrieval validation process, which expands on the simple comparison between the predictions and target described in the Model Development section. Although we focus primarily on validation of the L2 data, we also present validation statistics for L3 data in the Supplemental Material, which were nearly identical to the L2 results.

To assess performance, we compared our SM retrievals to observations from ground stations within the ISMN "sparse validation networks\cite{Galle2015, Cook2016b, Zreda2012, su2011tibetan, Smith2012, Schaefer2007, Ardoe2013, Leavesley2008, Bell2013, osenga2021community}" and the SMAP CVS\cite{chan_development_2018} for a separate time window than was used for training. We chose to use the time period August 1, 2018 to December 31, 2020, as our validation time window, as choosing a window past the training window would not have allowed for analysis against SMAP CVS. For context, we performed the same validation on the SMAP Enhanced L3 Product and the current official v1.0 CYGNSS SM product from UCAR (referenced earlier).

\subsection*{Qualitative Comparison with SMAP}

\begin{figure}[H]
    \centering
    \includegraphics[width=1\linewidth]{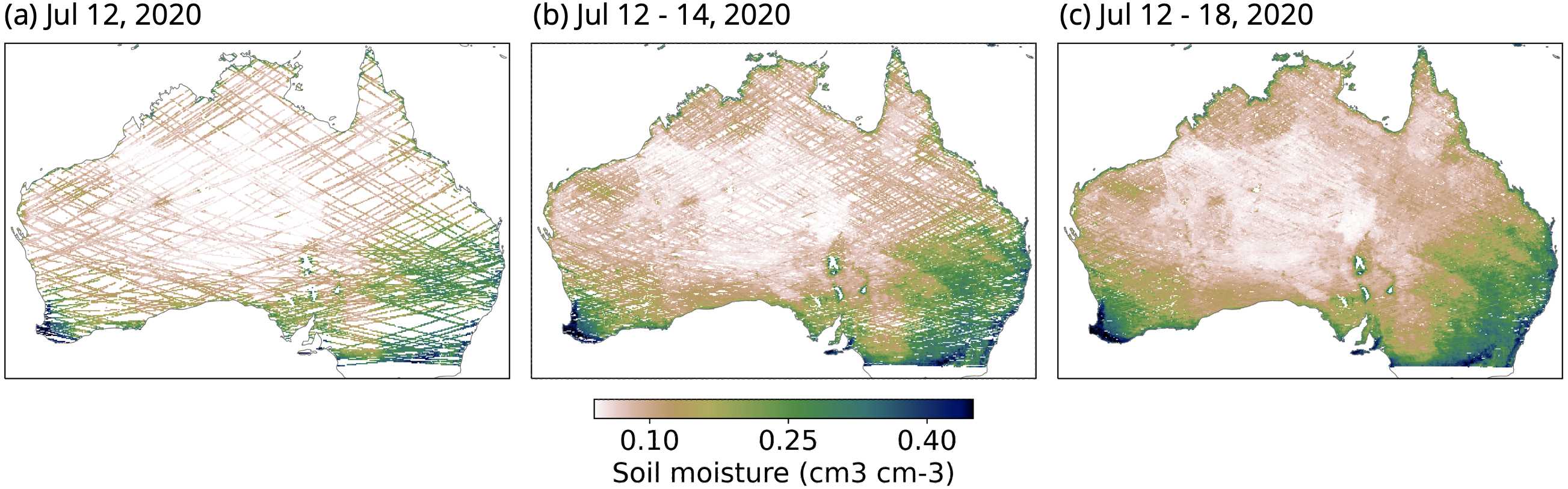}
    \caption{Example spatial coverage of the GNSS-R SM retrievals over 1, 3, and 7 days, when gridded to 9 km, over Australia.}
    \label{fig:Coverage_Australia}
\end{figure}

Evident in Figure \ref{fig:coverage}, the CYGNSS tracks do not provide complete coverage every day at the assumed spatial resolution. However, the SM retrievals can be gridded and averaged over select time periods to achieve complete spatial coverage. Figure \ref{fig:Coverage_Australia} shows example spatial coverage over Australia over a period of one (1), three (3), and seven (7) days. However, this coverage varies by latitude for CYGNSS, with equatorial latitudes having the sparsest coverage. There is an inherent trade off in the duration of this averaging window, i.e.,  the longer the window, the more complete the coverage. But with longer averaging windows, temporal dynamics are increasingly smoothed, though the longer-duration averaging allows for investigation of spatial patterns of SM.

\begin{figure}[H]
    \centering
    \includegraphics[width=1\linewidth]{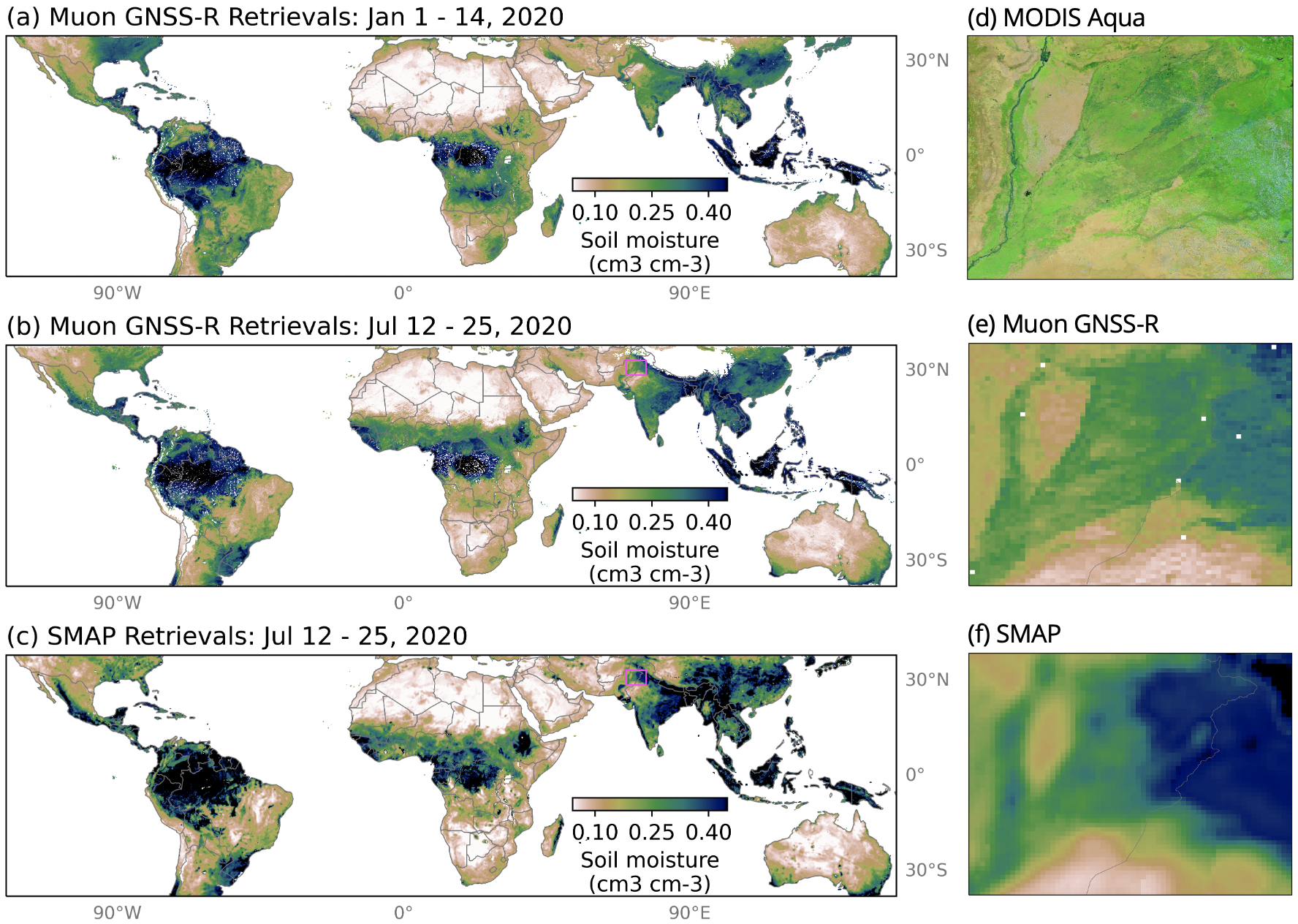}
    \caption{(a) Muon L3 9 km gridded retrievals of SM, averaged from Jan 1 - 14, 2020. Longitudinal bounds have been limited for clarity. (b) Same as (a), but for Jul 12 - 25, 2020. (c). Same as (b), but showing SM retrievals from the SMAP Level 3 Enhanced 9-km product. (d) A false color composite image from MODIS Aqua of the region outlined by the pink box in (b-c) for Jul 16 - 25, 2020, courtesy of NASA Worldview. (e) A zoomed in image of the region outlined by the pink box over India/Pakistan in (b). (f) A zoomed in image of the region outlined by the pink box over India/Pakistan in (c).
}
    \label{fig:Validation_GlobalFigure}
\end{figure}

In order to easily visualize differences in our SM product over time as well as differences between our SM product and other satellite SM products, we averaged our L3 retrievals over two-week time periods (Figure \ref{fig:Validation_GlobalFigure}). This figure shows an average of our SM retrievals for the first two weeks of January 2020 (Figure \ref{fig:Validation_GlobalFigure}a) and the second two weeks in July 2020 (Figure \ref{fig:Validation_GlobalFigure}b), highlighting the temporal variation observed. The eastern half of the United States, southern Africa, and northern Australia all show significant drying, whereas the Sahel and many parts of India show wetter conditions in July. For comparison, mean retrievals from the SMAP Enhanced product, which nominally has a repeat period of 2-3 days, are also shown for the same July time period (Figure \ref{fig:Validation_GlobalFigure}c). These figures show notable similarities and differences between the mean CYGNSS retrievals and those from SMAP. Regional spatial patterns agree, i.e., where SMAP shows SM as dry, so does CYGNSS, and vice versa. However, there are notable small-scale differences, which are more easily seen by the inset panels in Figure \ref{fig:Validation_GlobalFigure}d-f. These panels show a false-color image from MODIS over the Punjab region on the border of India and Pakistan (Figure \ref{fig:Validation_GlobalFigure}d), along with our CYGNSS GNSS-R SM retrievals (\ref{fig:Validation_GlobalFigure}e) and SMAP SM (\ref{fig:Validation_GlobalFigure}f). In the CYGNSS retrievals, there are sharper delineations between the deserts and agricultural regions than in the SMAP product. This is likely due to the fact that, although the SMAP Enhanced product is gridded to 9 km, its inherent sensor spatial resolution is 36 km, which would lead to greater spatial blurring relative to CYGNSS SM retrievals.

\subsection*{In Situ Validation}

As mentioned above, we validated our L2 and L3 SM retrievals against in situ SM data obtained from the ISMN and SMAP CVS. All in situ sites that met the following requirements were used for validation:

\begin{itemize}[noitemsep]
    \item SM data must be at 5 cm depth
    \item Location of site must be within +/-40 degrees latitude
    \item There must be at least 30 days of data during the validation time window (August 1, 2018 - December 31, 2020).
\end{itemize}

The vast majority of the in situ sites used for validation are located in the United States. A map of the stations used for validation is shown in Figure \ref{fig:coverage}.

\subsubsection*{Level 2 Retrievals}
The L2 SM retrievals are ungridded, trackwise measurements. In order to more directly compare with the official CYGNSS product from UCAR, which is a 36 km product, for our validation of the L2 data we upscaled our CYGNSS SM retrievals as well as the SMAP Enhanced product retrievals to also be at 36 km. This also allowed for a more direct comparison with the SMAP CVS data, which for most sites is posted at a similar resolution. In order to upscale our CYGNSS retrievals, we created daily averaged retrievals for all observations that fell within a 36 km diameter footprint centered on each in situ site location. In a similar manner, we created 36 km daily averaged SMAP retrievals by collecting all retrievals from 9 km cells within the same 36 km region around in situ sites. The daily averaged values were then directly compared with in situ SM observations that were also averaged at the same time interval. Note that there is still some discrepancy between this method of averaging and the spatial scale of the UCAR retrievals, as the UCAR retrievals are posted at a 36 km EASE-2.0 grid, the centers for which often do not exactly coincide with the location of in situ sites. For the UCAR product, we used retrievals from the grid cell center closest to each in situ site.

\begin{figure}[H]
    \centering
    \includegraphics[width=1\linewidth]{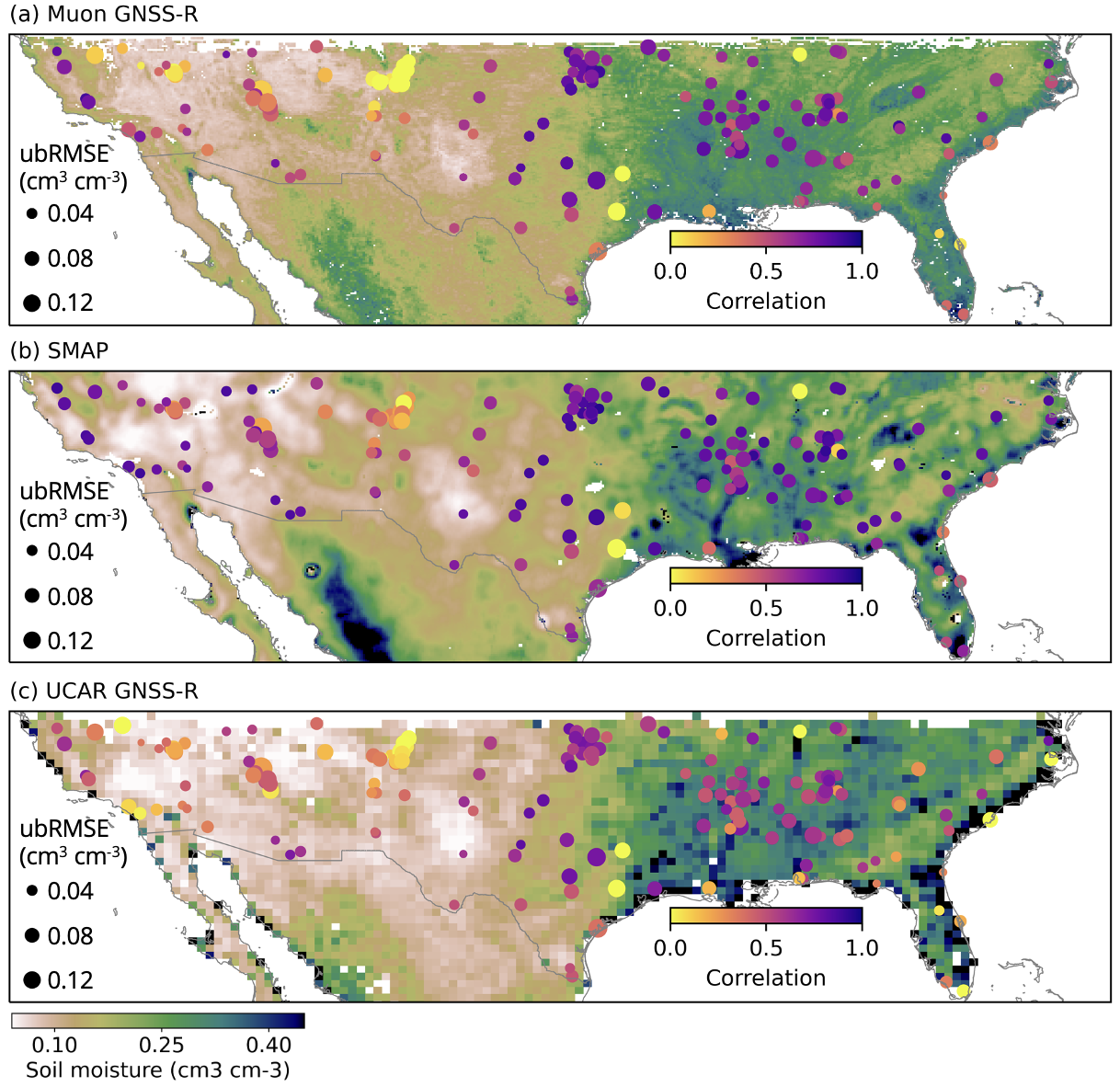}
    \caption{(a) Muon GNSS-R SM retrievals, gridded to 9 km, for the month of July 2020 (underlying map) with sparse network validation statistics of correlation and ubRMSE for the Level 2 retrievals (colored dots). (b) Same as (a), except for the SMAP Level 3 Enhanced Product, also gridded to 9 km. (c) Same as (a), except for the UCAR/CU SM product, which is gridded to 36 km.}
    \label{fig:Validation_Level2Dots}
\end{figure}

Standard validation statistics, including the unbiased root mean square error (ubRMSE), correlation coefficient, and bias were calculated for matchups at each site. Figure \ref{fig:Validation_Level2Dots} summarizes the correlation (colored dots) and ubRMSE (dot sizing) at each of the sparse network validation sites in the continental United States for the upscaled L2 Muon GNSS-R retrievals (Figure \ref{fig:Validation_Level2Dots}a), upscaled SMAP retrievals (Figure \ref{fig:Validation_Level2Dots}b), and the UCAR GNSS-R retrievals (Figure \ref{fig:Validation_Level2Dots}c). Broadly, spatial patterns of both correlation and ubRMSE are similar across all three products. Correlation tends to be higher at in situ sites in the eastern half of the United States relative to the western half, and sites that have low ubRMSE for one product are also likely to have low ubRMSE for all three products, though the values themselves are not exactly the same.

\begin{figure}[H]
    \centering
    \includegraphics[width=1\linewidth]{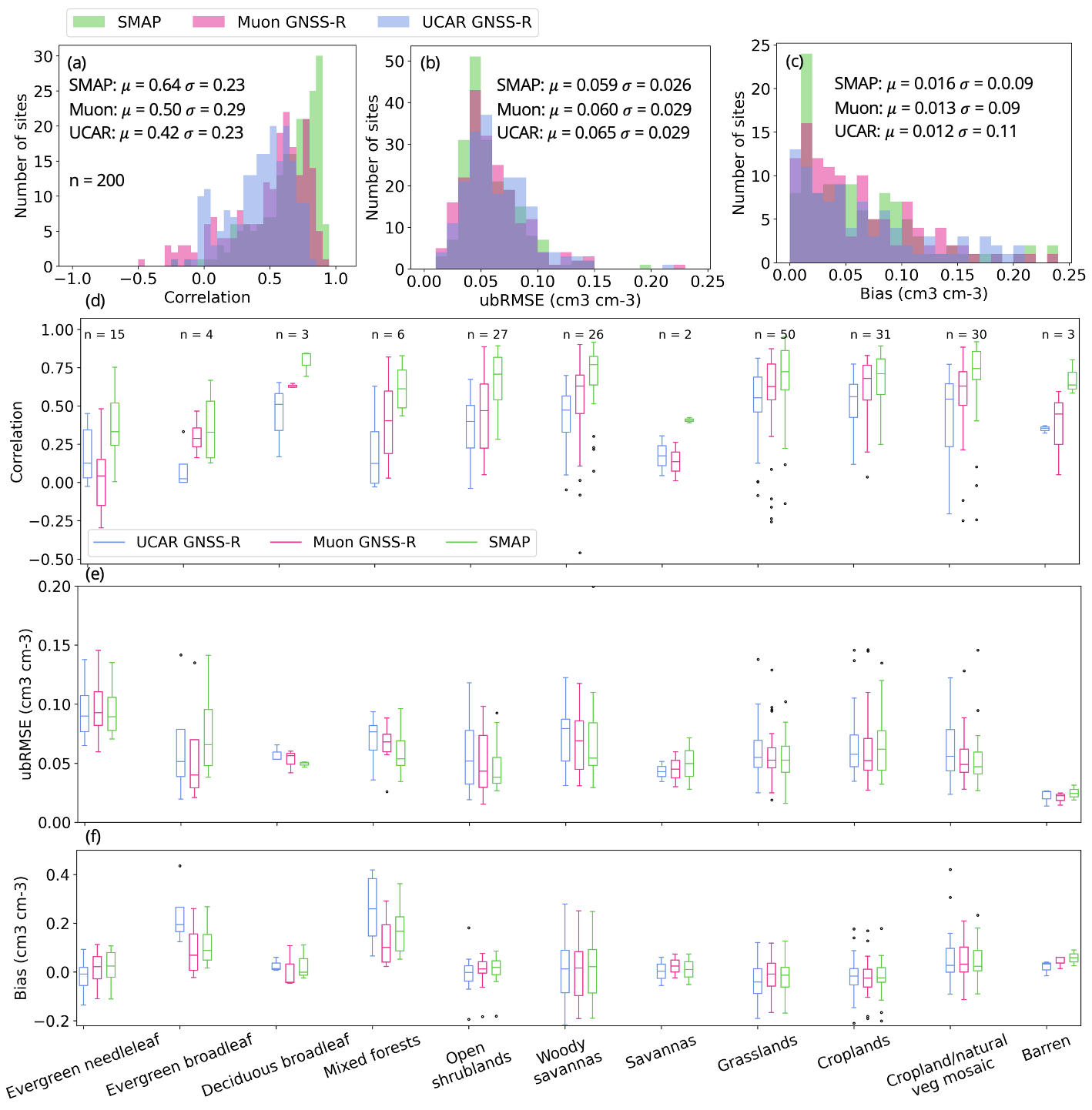}
    \caption{Aggregate validation statistics at sparse network validation sites for retrievals from SMAP (green), Muon GNSS-R (pink), and UCAR GNSS-R (blue), including correlation (a), ubRMSE (b), and bias (c). Correlation, ubRMSE, and bias are also split by land cover class in the box and whisker plots in (d-f), respectively. Land cover classes with n=1 were excluded from those shown in (d-f), though they are present in the aggregated histograms in (a-c).}
    \label{fig:Validation_Level2Landcover}
\end{figure}

Distributions of the correlation, ubRMSE, and bias for all three products at the sparse network validation sites are shown in Figure \ref{fig:Validation_Level2Landcover}a-c. Mean values of ubRMSE and bias are similar for all three products, though the distribution of the correlation coefficients are notably shifted lower for the UCAR GNSS-R retrievals than for the Muon GNSS-R retrievals or for SMAP. The Muon GNSS-R retrievals outperform the UCAR retrievals, with a mean increase in correlation of 0.08, though this is still 0.14 lower than the mean correlation for SMAP. The mean correlation coefficient for the Muon GNSS-R retrievals is significantly affected by the cluster of sites with a correlation near or below 0, whereas the peak of the distribution is much closer to that of SMAP than the mean value would suggest. Nearly all of the sites with extremely low correlation coefficients are located in forested, mountainous terrain, where the reflected signal strength is barely above the noise floor of the DDM. More stringent quality control measures could remove these degraded observations, though implementing these measures would also result in the removal of a portion of the retrievals in mountainous terrain. These are important regions for estimating SM in fire-prone areas - a focus of Muon's ongoing research - and will therefore be prioritized in near term model improvement efforts. 

The relative difficulty of retrieving SM in forests and/or mountainous terrain is shown in Figure \ref{fig:Validation_Level2Landcover}d-f, which splits the validation statistics for the sparse network sites by IGBP landcover class. Forests consistently show worse correlation and higher ubRMSE across all three products relative to lower vegetation environments like grasslands and croplands. In low- to moderately-vegetated environments, GNSS-R SM correlation and ubRMSE values approach those of SMAP.

\begin{figure}[H]
    \centering
    \includegraphics[width=1\linewidth]{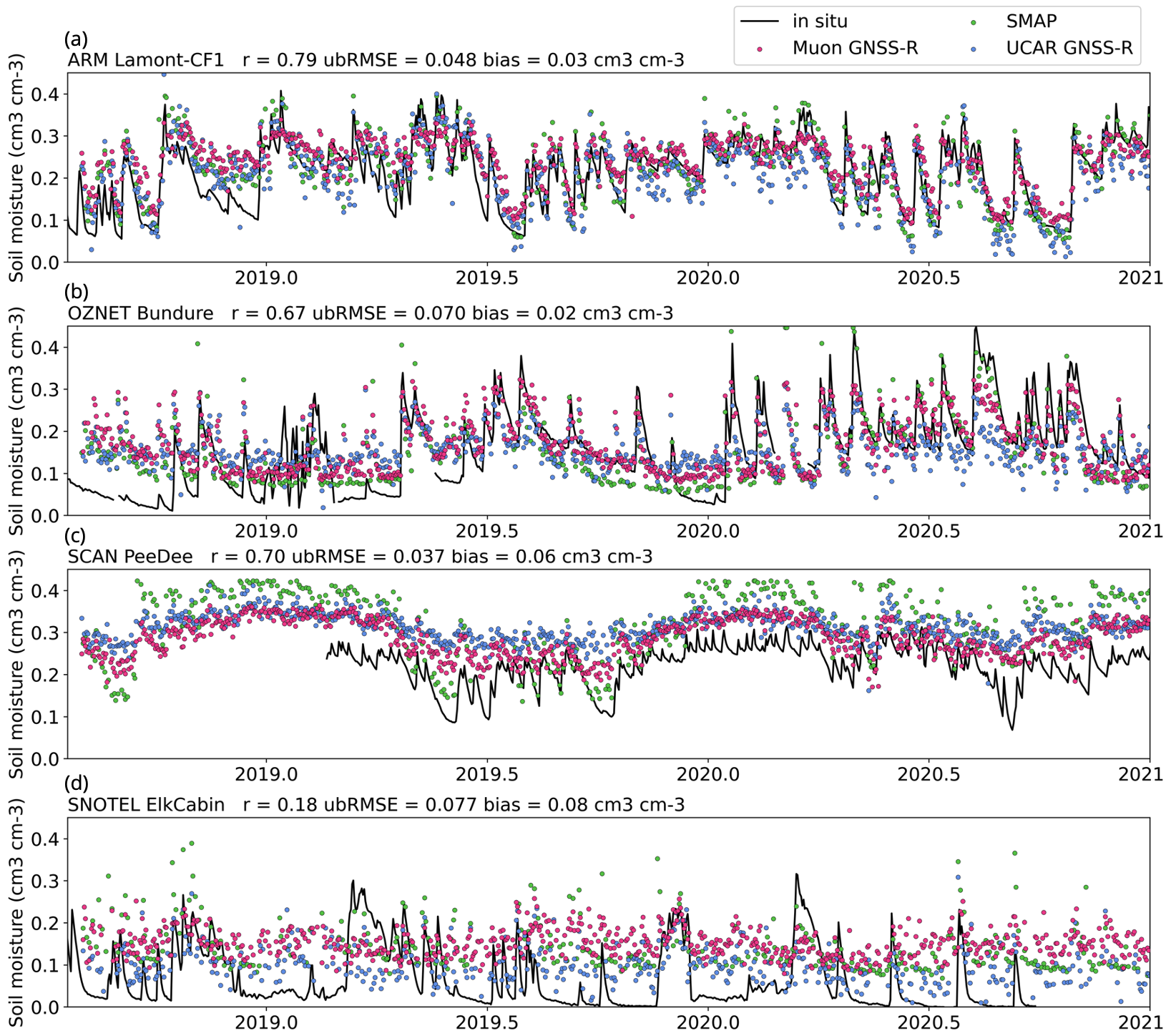}
    \caption{Time series of in situ SM (black lines) and SM retrievals from SMAP (green dots), Muon GNSS-R (pink dots), and UCAR GNSS-R (blue dots) from sparse network validation sites encompassing croplands (a), grasslands (b), woody savannas (c), and evergreen needleleaf forests (d). These sites were chosen because they had close to the Muon GNSS-R median correlation as found for sites within their land cover class (Figure \ref{fig:Validation_Level2Landcover}). The retrievals have not been bias corrected. Correlation, ubRMSE, and bias statistics for the Muon GNSS-R retrievals are shown at the top of each time series.}
    \label{fig:Validation_Level2TimeSeries}
\end{figure}

Figure \ref{fig:Validation_Level2TimeSeries} compares the upscaled Level 2 Muon GNSS-R retrievals (pink dots) against in situ observations (black lines) for four different land cover classes. These sites were chosen because the correlation coefficients between the Muon GNSS-R retrievals and the in situ observations were close to their median land cover class value. Retrievals from SMAP (green dots) and the UCAR GNSS-R product (blue dots) are also shown for context.

Using in situ SM observations, which are effectively point-scale measurements, to validate satellite remote sensing retrievals, which represent SM at a much larger spatial scale, is an imperfect exercise. Relative errors quantified using these “sparse” networks is very often higher than those quantified using sites with multiple probes spread across the expected spatial resolution of the satellite footprint, such as the SMAP CVS. This is both due to the discrepancy in the spatial sensing regions of SM probes versus satellite retrievals, as well as the fact that stations comprising sparse networks are often in more heterogeneous environments with denser vegetation than the SMAP CVS. Because of this, the ubRMSE value of 0.04 cm3 cm-3 that is considered to be the threshold for success for a satellite SM retrieval is less often achieved at sparse network sites than at the SMAP CVS.

Validation statistics for the upscaled L2 Muon GNSS-R retrievals at the SMAP CVS are shown in Table \ref{tab:Validation_TableCVSLevel2}. At the SMAP CVS, all three SM products have mean ubRMSEs below 0.04 cm3 cm-3 and very low bias. Notably, the mean correlation for the Muon GNSS-R retrievals is significantly higher (r = 0.72) than the mean value of the sparse network sites (r = 0.50). This is likely due to the larger number of probes being averaged such that the sites in the SMAP CVS are more representative of SM at the scale of the satellite footprint and because there are no forested sites within the sites in the SMAP CVS that are within the latitudinal range of CYGNSS.

\begin{table}[!htp]\centering
\small
\begin{tabular}{|M{0.09\linewidth}|M{0.07\linewidth}|M{0.07\linewidth}|M{0.07\linewidth}|M{0.07\linewidth}|M{0.07\linewidth}|M{0.07\linewidth}|M{0.07\linewidth}|M{0.07\linewidth}|M{0.07\linewidth}|M{0.07\linewidth}|}
\hline
\rowcolor{cyan}
&\multicolumn{3}{c}{\textbf{Correlation}} &\multicolumn{3}{c}{\textbf{ubRMSE (cm3 cm-3)}} &\multicolumn{3}{c}{\textbf{Bias (cm3 cm-3)}} \\
\textbf{Site name} &Muon GNSS-R &UCAR GNSS-R &SMAP &Muon GNSS-R &UCAR GNSS-R &SMAP &Muon GNSS-R &UCAR GNSS-R &SMAP \\
\hline
0701 Yanco &0.72 &0.66 &0.79 &0.043 &0.039 &0.045 &0.05 &0.04 &0.03 \\
\hline
1601 Walnut Gulch &0.35 &0.46 &0.71 &0.034 &0.033 &0.033 &0.03 &-0.02 &0.02 \\
\hline
1602 Little Washita &0.86 &0.64 &0.95 &0.023 &0.034 &0.02 &0.02 &-0.04 &0.01 \\
\hline
1603 Fort Cobb &0.84 &0.78 &0.9 &0.031 &0.036 &0.029 &-0.02 &-0.04 &-0.02 \\
\hline
1604 Little River &0.68 &0.37 &0.77 &0.032 &0.035 &0.046 &0.11 &0.1 &0.1 \\
\hline
1902 MonteBuey &0.74 &0.42 &0.84 &0.033 &0.046 &0.031 &-0.01 &-0.03 &-0.02 \\
\hline
4501 Niger &0.81 &0.66 &0.92 &0.026 &0.024 &0.021 &0.04 &0.01 &0.01 \\
\hline
4801 TxSON &0.81 &0.71 &0.93 &0.032 &0.038 &0.025 &-0.01 &-0.02 &0 \\
\hline
Mean of all &0.72 &0.59 &0.85 &0.032 &0.035 &0.031 &0.02 &0 &0.02 \\
\hline
\end{tabular}
\caption{Validation statistics for the L2 Muon GNSS-R retrievals, L3 UCAR/CU GNSS-R retrievals, and the SMAP Enhanced Product at the SMAP Core Validation sites within the latitudinal band of CYGNSS.}
\label{tab:Validation_TableCVSLevel2}
\end{table}

\subsubsection*{L3 Retrievals}
The L3 SM retrievals are already gridded to the same 9 km EASE-2.0 grid as the SMAP Enhanced Level 3 Product. We conducted the same validation analysis with the L3 retrievals as with the L2 retrievals; however, we did not upscale the L3 retrievals to 36 km as we did with the L2 analysis. Instead, we matched the daily L3 retrievals with the in situ sites that fell within each 9 km grid cell, using the same requirements for the in situ sites outlined at the beginning of this section. The SMAP Enhanced L3 Product was matched in the same way, and because the two products share the same grid, the same grid cells were matched with the same in situ sites for each product. Because the v1.0 UCAR GNSS-R product is not available at 9 km, we did not compare our L3 validation statistics with the UCAR product.

Validation statistics for the L3 product are very similar to those obtained for the L2 analysis. Because of this, we include the relevant figures in the Supplemental Material (Figures 4-6). Table 3 also shows the equivalent statistics for the L3 retrievals at the SMAP CVS. Because of the 9 km gridding scheme, the time series plots in Figure 6 are more sparse than those showing the upscaled, 36 km L2 retrievals in Figure \ref{fig:Validation_Level2TimeSeries}.

Overall, both the L2 and L3 retrievals perform similarly well at the sparse network validation sites and SMAP CVS, with mean ubRMSE values below 0.04 cm3 cm-3 at the SMAP CVS. The Muon GNSS-R retrievals outperform the official v1.0 CYGNSS product based on the UCAR/CU algorithm, with increased correlation coefficients and finer spatial resolution. Future refinements to the algorithm and increasing the receiver antenna gain on subsequent GNSS-R spacecraft could lead to better agreement with in situ sites in forested, mountainous environments.




\section*{Usage Notes}

Like most remote sensing products of soil moisture, the product presented here can be used by those interested in monitoring the evolution of drought or forecasting floods. Although the latency of the product is likely too long for assimilation into operational numerical weather models, this product could be used to retrospectively understand the influence of soil moisture on extreme weather events. The hourly gridded soil moisture files might be of particular interest to those researching applications related to the diurnal variability of soil moisture or its influence on the generation of convective storms. Finally, although the daily gridded soil moisture product contains spatial gaps at the 9 km scale, users who are interested in regional analyses of daily soil moisture could grid the Level 2 trackwise files to coarser grid cell sizes to achieve daily coverage in their region of interest, allowing for the estimation of soil moisture memory or research into any of the aforementioned applications.




\bibliography{GNSS-R_Nature_Data, GNSS-R_Nature_Data_extra}


\section*{Acknowledgements}

We acknowledge the use of imagery from the NASA Worldview application (https://worldview.earthdata.nasa.gov/), part of the NASA Earth Observing System Data and Information System (EOSDIS).

We acknowledge the work of Dennis D. Baldocchi and the FLUXNET-AMERIFLUX team, Frank Annor and Nicolaas Cornelis van de Giesen and the Trans-African Hydro-Meterological Observatory (TAHMO) network community, and the work of Ileen de Kat and Richard de Jeu as well as the VDS network team in support of the ISMN.

\section*{Author contributions statement}

M.R. and I.C. aggregated dataset, designed the data pipeline, and developed/tested DL models. C.C. developed the fractional land cover dataset. M.R., I.C., and C.C. analyzed the results and supported iteration on the model. C.C. formatted the retrievals into final data products. All authors reviewed the manuscript.



\end{document}